\documentclass[10pt,journal,compsoc]{IEEEtran}
\usepackage{multirow}
\usepackage{booktabs} % For formal tables
\usepackage[font=large,labelfont=bf]{caption}
\usepackage{subfigure}
\usepackage{tabularx}
\usepackage{multirow}
\usepackage{array}
\usepackage{amsfonts}
\usepackage{amsmath}
\usepackage{enumitem}
\usepackage{graphicx}
\usepackage{url}
\usepackage{dsfont}
\usepackage{enumerate}
\usepackage{times}
\usepackage{balance}
\usepackage{algorithm}
\usepackage{algorithmic}
\usepackage{hyperref}

\newlist{Properties}{enumerate}{2}
\setlist[Properties]{label=Property \arabic*.,itemindent=*}
\ifCLASSOPTIONcompsoc
  \usepackage[nocompress]{cite}
\else
  \usepackage{cite}
\fi
\ifCLASSINFOpdf
\else
\fi
\begin{document}
\title{Adversarial Attack on Large Scale Graph}
\author{Jintang Li, Tao Xie, Liang Chen*\thanks{*Corresponding author}, Fenfang Xie, Xiangnan He, Zibin Zheng}
\IEEEtitleabstractindextext{%
  \begin{abstract}
    Recent studies have shown that graph neural networks (GNNs) are vulnerable against perturbations due to lack of robustness and can therefore be easily fooled. Currently, most works on attacking GNNs are mainly using gradient information to guide the attack and achieve outstanding performance. However, the high complexity of time and space makes them unmanageable for large scale graphs and becomes the major bottleneck that prevents the practical usage. We argue that the main reason is that they have to use the whole graph for attacks, resulting in the increasing time and space complexity as the data scale grows. In this work, we propose an efficient \textbf{Simplified Gradient-based Attack (SGA)} method to bridge this gap. SGA can cause the GNNs to misclassify specific target nodes through a multi-stage attack framework, which needs only a much smaller subgraph. In addition, we present a practical metric named \textbf{Degree Assortativity Change (DAC)} to measure the impacts of adversarial attacks on graph data. We evaluate our attack method on four real-world graph networks by attacking several commonly used GNNs. The experimental results demonstrate that SGA can achieve significant time and memory efficiency improvements while maintaining competitive attack performance compared to state-of-art attack techniques. Codes are available via: \url{https://github.com/EdisonLeeeee/SGAttack}.
  \end{abstract}

  }

\maketitle

\IEEEdisplaynontitleabstractindextext

% \fi
\IEEEpeerreviewmaketitle

\IEEEraisesectionheading{\section{Introduction}}
\IEEEPARstart{R}{ecently}, with the enormous advancement of deep learning, many domains like speech recognition \cite{GravesMH13} and visual object recognition \cite{parkhi2015deep}, have achieved a dramatic improvement out of the state-of-the-art methods. Despite the great success, deep learning models have been proved vulnerable against perturbations. Specifically, Szegedy et al. \cite{goodfellow2014explaining} and Goodfellow et al. \cite{szegedy2013intriguing} have found that deep learning models may be easily fooled when a small perturbation (usually unnoticeable for humans) is applied to the images. The perturbed examples are also termed as ``adversarial examples'' \cite{szegedy2013intriguing}.

%Since our lives are surrounded by various graph networks, undoubtedly, graph-structured data plays a crucial role in many high-impact applications among the world\cite{ye2017efficient,ye2018identifying,liu2016community,kipf2016semi}. Therefore, there is no need to emphasize too much on the importance of the robustness of deep learning models on graph data. Nevertheless, it's regretful that adversarial examples are devastating in graph domains likewise and remain a major obstacle to overcome.

\begin{figure}[t]
  \centering
  \includegraphics[width=\linewidth]{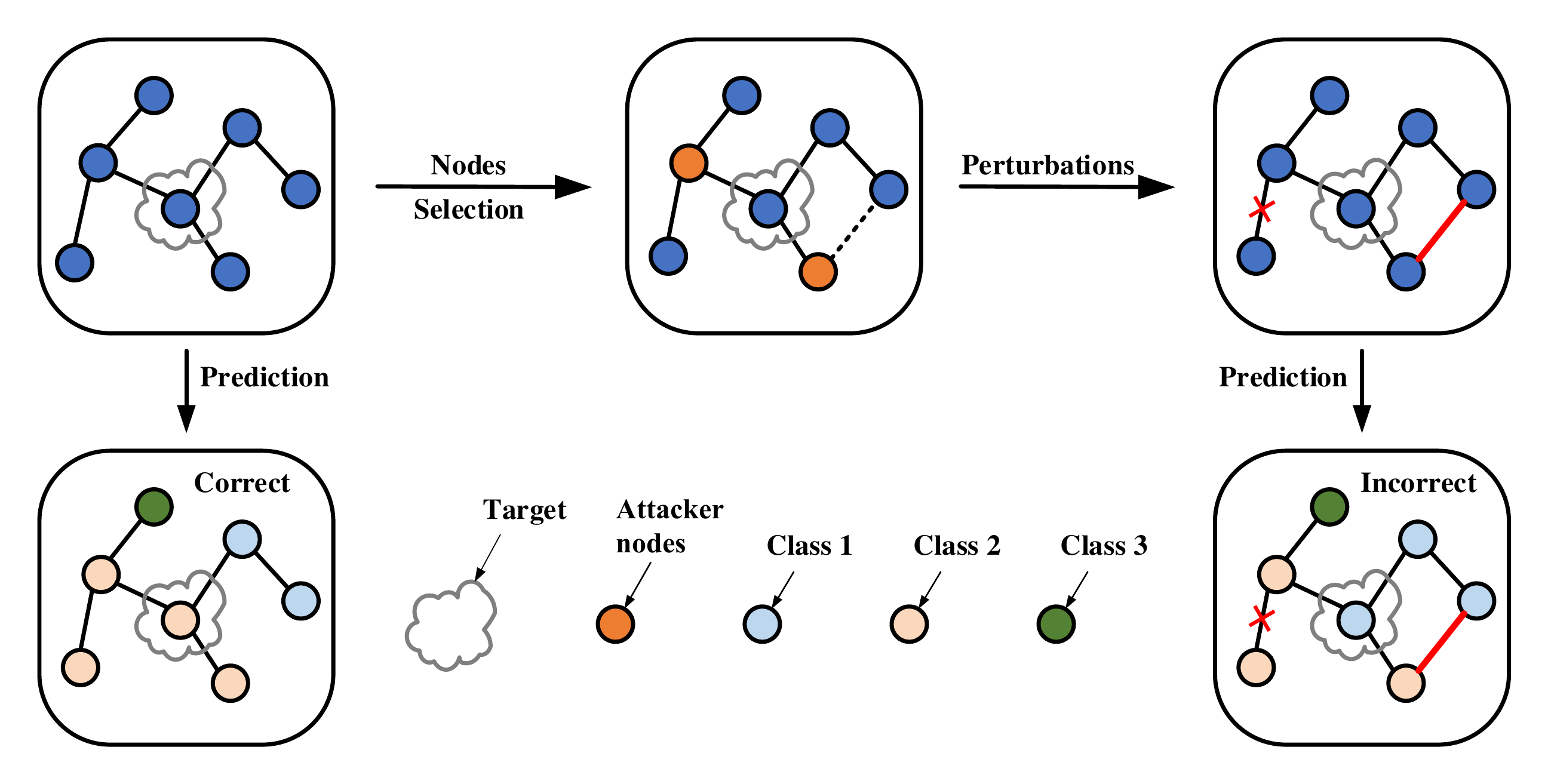}
  \caption{Adversarial attacks on the graph structure. Attackers tend to flip the edges of attacker nodes and lead to the misclassification of the target node.}
  \label{fig1}
\end{figure}

Graph structures are ubiquitous in nature and society, there is a great deal of research interest in studying graph data \cite{xie2018weighted,xie2018factorization,ChenL0GZ19,kipf2016semi,chen2018heterogeneous}. Undoubtedly, graph plays a crucial role in many high impact applications in the world \cite{ye2018identifying,liu2016community,kipf2016semi}. Therefore, the importance of the robustness of deep learning models on graph data should not be emphasized too much. However, the adversarial examples do have a significant effect on graph data, which is still a major obstacle to be overcome. So far, much of the current work on attacking graph neural networks has focused on the node classification task \cite{zugner2018adversarial, dai2018adversarial,bojchevski2019adversarial,zugner2019adversarial}. In order to fool a classifier and misclassify specific nodes, attackers may use two attack strategies to achieve the adversarial goals: \emph{direct attack} and \emph{influence attack}. For the direct attack, perturbations on the target node are allowed while the influence attack is limited to its neighboring nodes or beyond \cite{zugner2018adversarial}. Figure \ref{fig1} demonstrates a toy example of how deep learning models are fooled by attackers with small perturbations on the graph structure. In this case, the influence attack strategy is adopted.

In this paper, we focus on the \emph{targeted attack} that aims to make a specific node (e.g., a person in social networks \cite{tpds16}) misclassified. In this scenario,  Dai et al. \cite{dai2018adversarial} study the adversarial attack on graph structure data and propose a gradient-based method, namely GradArgmax, which modifies links based on gradients information of a surrogate model so as to fool the classifiers. In addition, Z\"{u}gner et al. \cite{zugner2018adversarial} propose Nettack, which is able to perturb the graph structure as well as node features. Despite great success, there are still some challenges for the attackers.

\textbf{Challenges}: (i) \emph{Scalability}. Most attacks have to store unnecessary graph information and thus suffer from the rising time and memory costs as the data scale increases. Naturally, they fail to efficiently conduct attacks on a larger graph. However, multi-million-scale graph networks are common in the real-world, so existing methods need to be improved and scaled to larger graphs. (ii) \emph{Evaluation}. A further challenge is to quantify the effects of adversarial attacks on graph data. As the graph data is unable and meaningless to be converted to continuous form, the impacts on the graph data are unsuitable to be measured with $\ell_2$-norm or $\ell_\infty$-norm \cite{madry2017towards}, which is different from that on image data. This makes the evaluation of the attack impacts a difficult problem to solve.

In this work, we attempt to tackle these challenges by our proposed methods. Specifically, our methods include two parts: (i) \emph{SGA framework}. We argue that it is unnecessary to use the whole graph to attack since attackers simply focus on misclassifying several nodes (the target nodes). Besides, due to the lack of time and space efficiency, Graph Convolution Network (GCN) \cite{kipf2016semi} is unsuitable as a surrogate model despite being used frequently in previous works. Inspired by Simplified Graph Convolutional Network (SGC) \cite{wu2019simplifying} and gradient-based attack methods \cite{dai2018adversarial,zugner2019adversarial}, we propose a novel \textbf{Simplified Gradient-based Attack (SGA)} framework for effective and efficient adversarial attacks. SGA only needs a much smaller subgraph consisting of $k$-hop neighbors of the target node ($k$ depends on the number of SGC layers and typically set to $2$), and sequentially flips edges with the largest magnitude of gradients in this subgraph by utilizing the surrogate model SGC. In addition, we introduce a scale factor to cope with the gradient vanishing during attacks (See Section \ref{sec:gradient}). Notably, our experimental evaluation suggests that the simplifications can hardly affect the attack performance. Moreover, the resulting model can remarkably improve the time and space efficiency --- even yields up to 1,976 and 5,753 times speedup over Nettack on Pubmed dataset in the direct attack and influence attack settings, respectively. Naturally, SGA can scale to very large graph networks easily. (ii) \emph{Degree Assortativity Change (DAC)}. Previous studies focus primarily on the intrinsic properties of graph \cite{newman2003mixing,foster2010edge}, but measuring the impact of graph adversarial attacks is however left unexplored so far. To address this problem, we also propose a practical metric named \textbf{Degree Assortativity Change (DAC)}, which can measure the intensity of the impact after perturbations being performed.

The main contributions of our works are summarized as follows:
\begin{itemize}
  % \vspace{-1mm}
  \item We propose a novel adversarial attack framework SGA, which extracts a much smaller subgraph centered at the target node, thereby addressing the difficulty of conducting attacks on a large scale graph.
  \item We notice the problem of gradient vanishing for gradient-based attack methods and solve it by introducing a scale factor to calibrate the model.
  \item We emphasize the importance of \emph{unnoticeability} for adversarial attacks on graph data and propose, first of all, a practical metric DAC to measure the attack impacts conveniently. This work can also be further developed in the graph domains.
  \item We conduct extensive experiments on four datasets by attacking several widely adopted graph neural networks. The experimental results show that our attack method has achieved significant improvements in both time and space efficiency while maintaining a significant attack performance compared to other state-of-the-art attack methods.
\end{itemize}

\section{Related Work}
Our work includes two parts: adversarial attack on graph data and evaluation of the attack impacts. In this section, we begin with the introduction of previous works of adversarial attacks on graph data and then discuss the details of ensuring the unnoticeability of adversarial attacks.

\subsection{Adversarial Attacks on Graph Data}
The robustness and security of deep learning models have received widespread attention, there has been a surge of interests in the adversarial attacks \cite{szegedy2013intriguing,goodfellow2014explaining,jagielski2018manipulating,ganju2018property,carlini2017towards}. The obtained results suggest that deep learning models are vulnerable to adversarial examples and could be easily fooled by them even under restricted black-box attack scenarios (attackers conduct attacks without any prior knowledge about the target model \cite{chen2020survey}).

% Deep learning models especially Graph Neural Networks (GNNs) have been proved extremely useful for node classification and other graph analysis tasks\cite{grover2016node2vec,perozzi2014deepwalk,kipf2016semi,wu2019simplifying,chiang2019cluster}. 

While previous works focus mostly on non-graph structure data (e.g., images and texts), adversarial attacks on graph structure data are considerably less studied because of the discreteness, which means that attackers must limit the attacks in order to maintain the graph property and allow only a few edges to be modified in the graph. To conduct a practical black-box attack on graph data with limited
knowledge, attackers often train a surrogate model locally to generate adversarial examples (perturbed graphs) and transfer them to fool the target models.

To be specific, we focus on the task of node classification in this work. By referring to \cite{chen2020survey}, we divide current approaches of adversarial attacks on the graph into two categories:

\subsubsection{Gradient-based attack}
This is the most commonly used method. Gradients have been successfully used to perform attacks in other domains \cite{goodfellow2014explaining,MoosaviDezfooli2016DeepFoolAS}. Since most of the existing models are optimized with gradient, along or against the direction of gradients is an efficient way to generate destructive adversarial examples. Focusing on the targeted attack, Dai et al. \cite{dai2018adversarial} propose GradArgmax, which extracts gradients of the surrogate model and deletes edges with the largest magnitude of the gradient to generate adversarial examples. However, the attacks are restricted to edge deletion only because the whole graph is stored with a sparse matrix (results in the lack of gradients information of non-edges). For the non-targeted attack, Z\"{u}gner et al. \cite{zugner2019adversarial} utilize the meta-gradients to solve the bi-level problem underlying the challenge of poisoning attacks (a.k.a, training-time attacks). Similarly, Xu et al. \cite{xu2019topology} propose PGD structure attack that conducts gradient attacks from a perspective of first-order optimization. Wang et al. \cite{WangLSLYZ20} introduce the approximate fast gradient sign method, which generates adversarial examples based on approximated gradients.

\subsubsection{Non-gradient-based attack}
As well as methods focused on gradients, attackers prefer to explore other heuristic algorithms to conduct attacks. Waniek et al. \cite{waniek2018hiding} propose ``Disconnect Internally, Connect Externally'' (DICE), conducting attacks by dropping edges between nodes with high correlations and connecting edges with low correlations.  Moreover, Z\"{u}gner et al. \cite{zugner2018adversarial} study both poisoning attacks and evasion attacks (a.k.a test-time attack) and further propose Nettack based on a linear GCN model, which maximizes the misclassification loss of the surrogate model greedily and perturbs the graph structure and node features to fool the classifiers. Zhang et al. \cite{zhang2020cross} introduce cross-entropy attack strategy based on the Deep Graph Infomax \cite{VelickovicFHLBH19} model. Chang et al. \cite{ChangRXHZC0H20} formulate the graph embedding method as a general graph signal process with corresponding graph filter and propose a generalized attacker under restrict black-box attack scenario.

\subsection{Unnoticeability of Adversarial Attacks}
Typically, in an adversarial attack scenario, attackers not only focus on the attack performance but also seek to be concealed to avoid detection. However, previous works usually conduct attacks under a fixed budget and thought it would be unnoticeable as if the budget is small enough. We argue that it is not sufficient to preserve the properties of graphs and ensure unnoticeability in most cases.

To bridge this gap, Z\"{u}gner et al. \cite{zugner2018adversarial} enforce the perturbations to ensure its \emph{unnoticeability} by preserving the graph's degree distributions and feature co-occurrences, restricting them to be marginally modified before and after attacks. However, there is still no practical metric to measure the impact of attacks on graph data.

On the other hand, researches on the graph (or network) structure have yielded several results with respect  to certain important properties, including ``small-world effect'' \cite{travers1977experimental,watts1998collective} and degree distributions \cite{barabasi1999emergence,amaral2000classes}. Unlike previous works, Newman et al. \cite{newman2003mixing} and Foster et al. \cite{foster2010edge} focus on another important network feature, i.e., \emph{assortativity}, and propose a number of measures of assortative mixing appropriate to various mixing types. Generally speaking, assortativity can be defined as the tendency of nodes to connect to each other, it is generally viewed as a metric to probe the properties of a specific graph, and also reflects the graph structure. Despite its popularity in network analysis and worthy of further study, it has not yet been used to measure the impacts of such adversarial attacks on graphs. Therefore, we aim to bridge the gap and apply it to graph adversarial learning.

\begin{table}[h]
  \centering
  \setlength{\belowcaptionskip}{10.0pt}
  \caption{Frequently used notations in this paper.}
  \label{notation}
  \resizebox{\linewidth}{!}{
    \begin{tabular}{c|c}
      \hline
      Notations                  &
      Descriptions                 \\

      \hline
      \hline
      \begin{tabular}[c]{@{}c@{}} $G$ \end{tabular}  &
      \begin{tabular}[c]{@{}c@{}} Graph representation of the data \end{tabular}    \\

      \hline
      \begin{tabular}[c]{@{}c@{}} $V$ \end{tabular}  &
      \begin{tabular}[c]{@{}c@{}} Set of vertices in the graph \end{tabular}    \\

      \hline
      \begin{tabular}[c]{@{}c@{}} $E$ \end{tabular}  &
      \begin{tabular}[c]{@{}c@{}} Set of edges in the graph \end{tabular}   \\

      \hline
      \begin{tabular}[c]{@{}c@{}} $\mathcal{C}$ \end{tabular} &
      \begin{tabular}[c]{@{}c@{}} Set of class labels of nodes \end{tabular}   \\

      \hline
      \begin{tabular}[c]{@{}c@{}} $N$, $C$, $F$ \end{tabular} &
      \begin{tabular}[c]{@{}c@{}} Number of nodes, classes and feature dimensions \end{tabular}   \\

      \hline
      \begin{tabular}[c]{@{}c@{}}  $A$ \end{tabular} &
      \begin{tabular}[c]{@{}c@{}} Adjacency matrix of the graph, $N\times N$ \end{tabular}
      \\

      \hline
      \begin{tabular}[c]{@{}c@{}}  $X$ \end{tabular} &
      \begin{tabular}[c]{@{}c@{}} Feature matrix of the nodes, $N\times F$ \end{tabular}
      \\

      \hline
      \begin{tabular}[c]{@{}c@{}}  $D$ \end{tabular} &
      \begin{tabular}[c]{@{}c@{}} Diagonal matrix of the degree of each vertex, $N\times N$ \end{tabular}
      \\

      \hline
      \begin{tabular}[c]{@{}c@{}}  $f_\theta$ \end{tabular} &
      \begin{tabular}[c]{@{}c@{}} Graph neural networks model \end{tabular}
      \\

      \hline
      \begin{tabular}[c]{@{}c@{}}  $W$ \end{tabular} &
      \begin{tabular}[c]{@{}c@{}} Trainable weight matrix, $F_l \times F_{l+1}$ \end{tabular}
      \\

      \hline
      \begin{tabular}[c]{@{}c@{}}  $Z$ \end{tabular} &
      \begin{tabular}[c]{@{}c@{}} Prediction probability of nodes, $N \times C$ \end{tabular}
      \\

      \hline
      \begin{tabular}[c]{@{}c@{}}  $t$ \end{tabular} &
      \begin{tabular}[c]{@{}c@{}} Target node to attack \end{tabular}
      \\

      \hline
      \begin{tabular}[c]{@{}c@{}}  $\epsilon$ \end{tabular} &
      \begin{tabular}[c]{@{}c@{}} Scale factor to calibrate the model \end{tabular}
      \\

      \hline
      \begin{tabular}[c]{@{}c@{}}  $\mathcal{D}(u,v)$ \end{tabular} &
      \begin{tabular}[c]{@{}c@{}} The shortest distance between $u$ and $v$ \end{tabular}
      \\

      \hline
      \begin{tabular}[c]{@{}c@{}}  $\mathcal{N}(t)$ \end{tabular} &
      \begin{tabular}[c]{@{}c@{}} Set of nodes adjacent to node $t$ \end{tabular}
      \\

      \hline
      \begin{tabular}[c]{@{}c@{}}  $k$ \end{tabular} &
      \begin{tabular}[c]{@{}c@{}} Radius of the subgraph \end{tabular}
      \\

      \hline
      \begin{tabular}[c]{@{}c@{}}  $\mathcal{A}$ \end{tabular} &
      \begin{tabular}[c]{@{}c@{}} Set of attacker nodes,\\ $\mathcal{A}=\{t\}$ for direct attack, \\ $\mathcal{A}=\mathcal{N}(t)$ for influence attack \end{tabular}
      \\

      \hline
      \begin{tabular}[c]{@{}c@{}}  $M$ \end{tabular} &
      \begin{tabular}[c]{@{}c@{}} Degree mixing matrix \end{tabular}
      \\

      \hline
      \begin{tabular}[c]{@{}c@{}}  $r$ \end{tabular} &
      \begin{tabular}[c]{@{}c@{}} Degree assortativity coefficient \end{tabular}
      \\

      \hline
      \begin{tabular}[c]{@{}c@{}}  $\Delta$ \end{tabular} &
      \begin{tabular}[c]{@{}c@{}} Attack budget \end{tabular}
      \\

      \hline
      \begin{tabular}[c]{@{}c@{}}  $G^\prime$, $A^\prime$, $X^\prime$ \end{tabular} &
      \begin{tabular}[c]{@{}c@{}} Perturbed graph, adjacency matrix and feature matrix \end{tabular}
      \\

      \hline
      \begin{tabular}[c]{@{}c@{}}  $\Tilde{\mathcal{L}}_t$, $\Tilde{\mathcal{L}}^{(sub)}_t$ \end{tabular} &
      \begin{tabular}[c]{@{}c@{}} Targeted misclassification loss on the graph and subgraph \end{tabular}
      \\

      \hline
      \begin{tabular}[c]{@{}c@{}}  $S$ \end{tabular} &
      \begin{tabular}[c]{@{}c@{}} Structure score matrix \end{tabular}
      \\

      \hline
      \begin{tabular}[c]{@{}c@{}}  $G^{(sub)}$ \end{tabular} &
      \begin{tabular}[c]{@{}c@{}} The $k$-hop subgraph \end{tabular}
      \\

      \hline
      \begin{tabular}[c]{@{}c@{}}  $V^{(sub)}$, $E^{(sub)}$ \end{tabular} &
      \begin{tabular}[c]{@{}c@{}} Set of nodes, edges in the subgraph \end{tabular}
      \\

      \hline
      \begin{tabular}[c]{@{}c@{}}  $A^{(sub)}$ \end{tabular} &
      \begin{tabular}[c]{@{}c@{}} Adjacency matrix of the subgraph \end{tabular}
      \\

      \hline
      \begin{tabular}[c]{@{}c@{}}  $E^{(exp)}$ \end{tabular} &
      \begin{tabular}[c]{@{}c@{}} Set of expanded edges in the subgraph \end{tabular}
      \\
      \hline
    \end{tabular}
  }
\end{table}

\section{PRELIMINARY}
Before presenting our proposed methods, we will first give the notations of graph data formally, and then introduce the surrogate models --- the GCN family, finally clarify the details of other proposed adversarial attack methods. See Table \ref{notation} for frequently used notations.

\subsection{Notations}
\label{sec3}
Specifically, we focus on the task of semi-supervised node classification in a single, undirected, attributed graph. Formally, we define $G=(A,X)$ as an attributed graph, where $A \in \{0, 1\} ^ {N\times N}$ is a symmetric adjacency matrix denoting the connections of the $N$ nodes, and $X \in \{0, 1\} ^ {N\times F}$ or $X \in \mathbb{R} ^ {N\times F}$ represent the binary or continuous node features with $F$ dimension. We use $V$ to denote the set of nodes and $E \subseteq V\times V$ the \emph{connected} edges of the graph $G$; $\mathcal{C}=\{c_i \}$ denotes a set of class labels where $c_i$ indicates the ground-truth label of node $i$, and we define $C$ as the number of classes in $\mathcal{C}$.

\subsection{Graph Convolution Network Family}
\label{gcn}
\subsubsection{Vanilla Graph Convolution Network (GCN)}
Since a number of existing works \cite{zugner2018adversarial,zugner2019adversarial,dai2018adversarial,xu2019topology} use vanilla GCN \cite{kipf2016semi} as a surrogate model to conduct attacks, thus we first introduce GCN and further draw attention on SGC \cite{wu2019simplifying} --- a simplified variant of GCN. Refer to this work \cite{kipf2016semi}, GCN is recursively defined as
\begin{equation}
  H^{(l+1)}=\sigma (\Tilde{D}^{-\frac{1}{2}}\Tilde{A}\Tilde{D}^{-\frac{1}{2}}H^{(l)}W^{(l)})\,,
\end{equation}
where $\Tilde{A}=A+I_N$ is the adjacency matrix with self-loops. $I_N$ is a $N$ by $N$ identity matrix, and $\Tilde{D}_{ii}=\sum_{j}\Tilde{A}_{ij}$ is a diagonal degree matrix. $W^{(l)} \in \mathbb{R}^{F_l \times F_{l+1}}$ is a trainable input-to-hidden weight matrix  and $H^{(l)} \in \mathbb{R}^{N \times F_l}$ is the matrix of hidden representation (activation), both of which are related to the $l^{th}$ layer. Particularly, $F_0=F$, $H^{(0)}=X$ is the input of neural network. $\sigma{(\cdot)}$ represents the element-wise activation function of network and is usually defined as $\mathrm{ReLU}\,(\cdot)=\max{(\cdot,0)}$.

For node classification task, consider GCN with one hidden layer, and let $\hat{A}=\Tilde{D}^{-\frac{1}{2}}\Tilde{A}\Tilde{D}^{-\frac{1}{2}}$, then the forward of GCN with pre-processing step could be taken as
\begin{equation}
  Z = f_\theta (A,X)=\mathrm{softmax}\,(\hat{A}\ \mathrm{ReLU}\,(\hat{A}XW^{(0)})\ W^{(1)})\,,
\end{equation}
% where the $\mathrm{softmax}$ function is defined as:
% \begin{gather*}
% 	\mathrm{softmax\,(Z_i)}=\frac{\mathrm{exp}\,(Z_i)}{\sum_{i}{\mathrm{exp}\,(Z_i)}}\,,
% \end{gather*}
where $Z \in \mathbb{R}^{N \times C}$ is the output matrix of GCN and indicates the prediction probability of nodes belonging to different classes. The softmax activation function, defined as $\mathrm{softmax}(x_i)=\frac{1}{z} \mbox{exp}(x_i)\,$with $z=\sum_j \mbox{exp}(x_j)$, is applied row-wise.

Given a set of labeled nodes $V_L \subseteq V$ with ground-truth labels $\mathcal{C}_L \subseteq \mathcal{C}$, the goal of GCN is to learn a mapping function $g: V \to \mathcal{C}$ by minimizing the cross-entropy loss:
\begin{equation}
  \label{gcnloss}
  \mathcal{L}(\theta;A,X) = -\sum _{i\in V_L }\ln{Z_{i,c_i}}\,,\quad Z=f_\theta (A,X)\,,
\end{equation}
where $c_i\in \mathcal{C}_L$ is the class label of node $i$ and $\theta=\{W^{(0)},W^{(1)} \}$ denotes the trainable weights of model. After being optimized with Gradient Descent \cite{bottou2010large}, the weights are learned to predict nodes in an unlabeled set $V_U=V-V_L$.

\subsubsection{Simplified Graph Convolutional Network (SGC)}
\textbf{Linearization}. The drawback of vanilla GCN is the excessive computational cost of message aggregation between nodes and their neighboring nodes, which is repeatedly and unnecessarily computed during training. To address this problem, Wu et al. \cite{wu2019simplifying} theoretically analyze the structure of GCN and further propose a linear variant, namely SGC. SGC replaces the nonlinear activation function $\mathrm{ReLU}\,(\cdot)$ with identity function and collapses weight matrices between consecutive layers. In this way, the forward of GCN can be simplified as
\begin{equation}
  \begin{split}
    \label{sgc}
    Z = f_\theta (A,X)&=\mathrm{softmax}\,(\hat{A}\ \mathrm{ReLU}\,(\hat{A}XW^{(0)})\ W^{(1)})\\
    &=\mathrm{softmax}\,(\hat{A}\hat{A}XW^{(0)}\ W^{(1)})\\
    &=\mathrm{softmax}\,(\hat{A}^2XW)\,,
  \end{split}
\end{equation}
where $W = W^{(0)}\ W^{(1)}$ is a collapsed weight matrix. With pre-computing $\hat{S}=\hat{A}^2X$, the complicated GCN structure can be simplified as an input-to-output fully-connected neural network \cite{rosenblatt1961principles} without any hidden units, thereby avoiding redundant computation and greatly reducing training time and memory.

\subsection{SOTA Adversarial Targeted Attack Methods}
For adversarial attacks on graphs, attackers may conduct \emph{structure attack} or \emph{feature attack} \cite{chen2020survey} to modify the graph structure or node features, respectively. Following the work \cite{zugner2018adversarial}, we assume that attackers have prior knowledge about the graph structure and node features, including ground-truth labels of nodes. Beyond that, attackers are not allowed to access any additional information about the target models, neither model architecture nor parameters. In this case, attackers often train a transferable surrogate model locally, perform perturbations to fool it to achieve the best result of misclassification, and eventually transfer to other target models. Since we focus on the targeted attack in node classification task, here we briefly introduce other proposed state-of-the-art targeted attack methods: \textbf{Nettack} \cite{zugner2018adversarial} and \textbf{GradArgmax} \cite{dai2018adversarial}.

\subsubsection{Nettack}
Similar to SGC, Nettack uses a linear variant of two-layer GCN as a surrogate model to conduct targeted attack. Given a target node $t$ and a budget $\Delta \in \mathbb{N}$, Nettack modifies the graph structure and node features aiming to maximize the misclassification loss of the surrogate model:

\begin{equation}
  \begin{aligned}
     & \arg \max_{A^\prime,X^\prime} (\max_{c^\prime_t \neq c_t} \ln{Z_{t,c^\prime_t}} - \ln{Z_{t,c_t}}),\quad t\in V
    \\
     & s.t. \sum_i \sum_j  |X_{i,j}-X^\prime_{i,j}  | + \sum_{u<v} |A_{u,v}-A^\prime_{u,v} | \leq \Delta
  \end{aligned}
\end{equation}
where $Z=f_\theta (A^\prime,X^\prime)$ is the output of the surrogate model, $A^\prime$ and $X^\prime$ are the perturbed adjacency matrix and feature matrix, respectively.

\subsubsection{GradArgmax}
GradArgmax uses vanilla GCN as a surrogate model to maximize the cross-entropy loss described in Eq.(\ref{gcnloss}) by gradient ascent. As the loss function $\mathcal{L}$ and target node $t$ are specified, GradArgmax computes the partial derivative of $\mathcal{L}_t$ with respect to each \emph{connected} edge of the adjacency matrix:
\begin{equation}
  \nabla _G = \nabla _{A} \mathcal{L}_t = \frac{\partial \mathcal{L}_t}{\partial A},
\end{equation}
where $\mathcal{L}_t$ denotes the targeted loss w.r.t node $t$.

To preserve the discreteness of adjacency matrix $A$, GradArgmax greedily modifies those edges with $\Delta$-largest magnitude of gradients. In addition, the adjacency matrix is stored as a sparse one in order to avoid excessive computational costs, but only gradients of the \emph{connected} edges are computed, which means that attacks are restricted to the edge deletion only, and the information on surrogate gradients is not fully utilized.

\section{SIMPLIFIED GRADIENT-BASED ATTACK}
In this work, we simply focus on the structure attack. We follow the attack settings of Nettack \cite{zugner2018adversarial} which modifies graph structure and node features by flipping them from 0 to 1 or vice verse. Since node features in the real-world dataset may be continuous (not binary), it is difficult to constrain attacks within a given budget $\Delta \in\mathbb{N}$. Note that, our method can be easily extended to the feature attack (either binary or continuous features) as well, just by taking into account the gradients of the input features. In the node classification scenario detailed in Section \ref{sec3}, the goal of an attacker is to perturb the original graph $G=(A,X)$ with limited budgets $\Delta$ and further lead to a misclassification of target models.

% Firstly, we introduce the set $\mathcal{A} \subseteq V$, denoting the influencer nodes in the graph. Attackers aim to attack target node $t$ by flipping a few edges $e\in \mathcal{A} \times V_A$ directly, where $V_A \subseteq V$ denotes the victim nodes. Particularly, $t \in V_A$ for direct attack, in contrast to $t \notin V_A$ for influence attack. To achieve more influence, usually $\mathcal{A} \subseteq \mathcal{N}(t)$, where $\mathcal{N}(t)$ is a set of nodes adjacent to $t$. As shown in Figure \ref{fig1}, the influencer nodes (orange ones) is selected from the neighborhood of target node.

To conduct attacks on the target models without additional prior knowledge, inspired by the adversarial methods mentioned above, the proposed SGA follows four steps: (i) Train a surrogate model locally (Section \ref{sec:surrogate}). (ii) Extract a $k$-hop  subgraph  (Section \ref{sec:subgraph}). (iii) Compute surrogate gradients via subgraph (Section \ref{sec:gradient}). (iv)  Choose to add or remove edges iteratively based on gradients (Section \ref{sec:update}).

\subsection{Surrogate Model}
\label{sec:surrogate}
The most widely used surrogate model is vanilla GCN, however, it has significant drawbacks as described in Section \ref{gcn}. We use a $k$-layer SGC as our surrogate model instead. SGC removes the nonlinear transition functions (e.g., ReLU) between each layer and only keep the final softmax, hence the output of a $k$-layer SGC can be formulated as
\begin{equation}
  \begin{split}
    Z = f_\theta (A,X)&=\mathrm{softmax}(\hat{A}\dots \hat{A} \hat{A}X W^{(0)}W^{(1)}\dots W^{(k-1)})\\
    &=\mathrm{softmax}(\hat{A}^{k}XW)
  \end{split}
\end{equation}
where $W=W^{(0)}W^{(1)}\dots W^{(k-1)}$ is the collapsed weight matrix.

First and foremost, we train SGC on the input graph until convergence with fine-tuned hyper-parameters. In fact, the goal is to obtain the weight matrix $\theta=\{W\}$, which will be used later to guide the attack in our method.

\subsection{$k$-hop Subgraph}
\label{sec:subgraph}
Given a target node $t$ with ground-truth class label $c_t$, the goal of attackers is to get it classified as another class $c^\prime$ where $c^\prime \neq c_t $.  Therefore, we design the targeted misclassification loss as follows:
\begin{equation}
  \label{sgaloss}
  \Tilde{\mathcal{L}}_t=\max_{c^\prime \neq c_t} \ln{Z_{t,c^\prime}} - \ln{Z_{t,c_t}},\quad t\in V\,,
\end{equation}
where $Z_{t,c_t}$ indicates the predicted probability that node $t$ belongs to class $c_t$. According to Eq.(\ref{sgaloss}), to compute the targeted loss $\Tilde{\mathcal{L}}_t$, we  only need to compute $Z_t$, a row-vector of $Z$. Therefore, computing $\{Z_{u}|\,u \in V\ \mbox{and} \ u \neq t\}$ is unnecessary and redundant. This motivates us to simplify the computation.

\begin{figure*}[t]
  \centering
  \includegraphics[width=\linewidth]{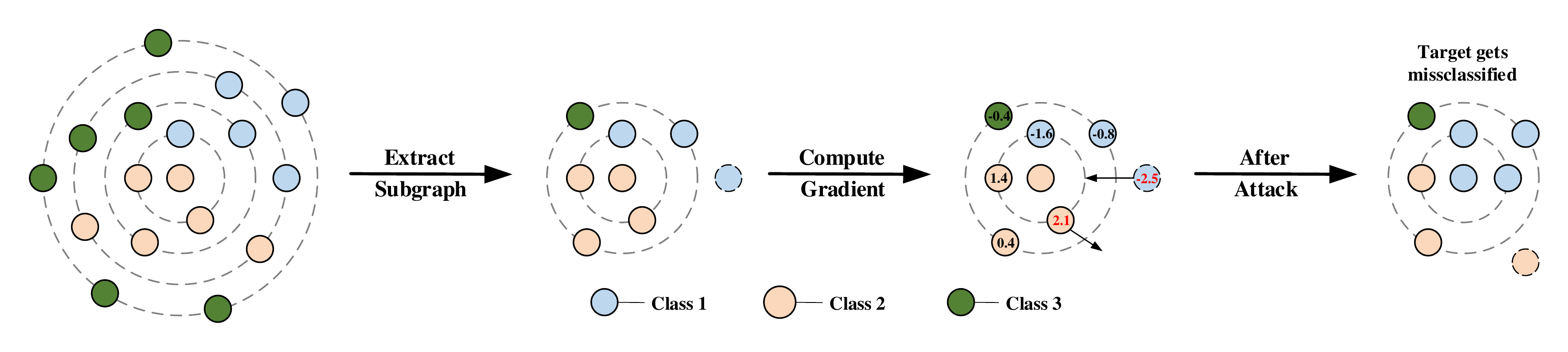}
  \caption{Simplified gradient-based attack via subgraph. A two-hop subgraph centered at target node is extracted and some potential nodes (dotted-line nodes) outside the subgraph are added as well to enlarge  the  possible  perturbation  set. The values within nodes denote the gradients.}
  \label{SGA}
\end{figure*}

\begin{Properties}
  \label{theorem}
  \item Given a normalized adjacency matrix $\hat {A}$ with self-loops, i.e., $\hat {A}_{u,u} \neq 0, \forall u \in V$. let $\mathcal{D}{(u,v)}$ denotes the shortest distance between $u$ and $v$, thus
  \begin{equation}
    \label{theoremEq}
    [\hat{A}^k]_{u,v}\begin{cases}
      =0,     & \mbox{if }\  \mathcal{D}(u,v)> k    \\
      \neq 0, & \mbox{if }\  \mathcal{D}(u,v)\leq k
    \end{cases}
  \end{equation}
\end{Properties}
% \begin{proof}
%   The detailed proof is omitted for the sake of brevity.
% \end{proof}

Given a specific target node $t \in V$, it is clear that $\Tilde{\mathcal{L}}_t$ depends on $Z_t$ only, where $Z_t=[\hat{A}^k]_tXW$. So what we need is to compute $[\hat{A}^k]_t$, a row-vector of $\hat{A}^k$. According to Property 1, we can construct a $k$-hop subgraph $G^{(sub)}=(A^{(sub)},X)$ consisting of:
\begin{equation}
  \begin{aligned}
     & V_s=\{u\,|\,\mathcal{D}(t,u)\leq k\},                                          \\
     & E_s=\{(u,v)\,|\,\mathcal{D}(t,u)\leq k\  \mbox{and}\  \mathcal{D}(t,v)\leq k\}
  \end{aligned}
\end{equation}
The set of nodes $V_s$ is also considered as the $k$-hop neighbors of target node $t$.

There is a crucial problem that only the gradients of \emph{connected} edge $e_i \in E_s$ will be computed, resulting in edge deletion only. To enlarge the possible perturbation set, it is also important to consider the gradients of the non-edges that might be connected later. Unfortunately, the possible number of non-edges is nearly $N^2$ and this makes the computation of gradients very costly. Inspired by DICE \cite{cai2005mining}, a straightforward way to attack is ``Disconnect Internally, Connect Externally''. Following this insight, we will only consider those nodes that belong to different classes to construct a non-edge set.

However, if a graph is large, there are still a great number of nodes that belong to different classes. To avoid excessive computation, we empirically add nodes and edges to fulfill
\begin{equation}
  \begin{aligned}
    \label{potiential}
    V_p & = \{u\,|\, c_{u}=c^\prime_t, u \in V\}\,, \\
    E_p & =\mathcal{A} \times V_p\,,
  \end{aligned}
\end{equation}
where $c^\prime_t =\arg \max_{c^\prime\neq c_t} Z_{t,c^\prime}$ is the next most probable class obtained from surrogate model SGC, $\mathcal{A} \subseteq V$ is the set of \emph{attacker nodes} and the perturbations are constrained to these nodes \cite{zugner2018adversarial}. In particular, we set $\mathcal{A}=\{t\}$ for direct attack and $\mathcal{A}=\mathcal{N}(t)$ for influence attack where $\mathcal{N}(t)$ is the set of neighboring nodes adjacent to $t$. We term these nodes in $V_p$ as ``potential nodes'' and edges in $E_p$  as ``potential edges'' since they may be included in this subgraph later. Eq.(\ref{potiential}) shows that we prefer to connect attacker nodes in $\mathcal{A}$ with potential nodes to influence the target node $t$. Therefore, the targeted loss can be also simplified as
\begin{equation}
  \label{final_loss}
  \Tilde{\mathcal{L}}_t^{(sub)}= \ln{Z_{t,c^\prime_t}^{(sub)}} - \ln{Z_{t,c_t}^{(sub)}}\,,\quad t\in V
\end{equation}
where $\Tilde{\mathcal{L}}_t^{(sub)}$ is derived from Eq.(\ref{sgaloss}) which denotes the targeted loss computed via subgraph, $Z_t^{(sub)} = f(A^{(sub)},X) \in \mathbb{R}^{1\times C}$ denotes the prediction probability vector of target node $t$ via subgraph.

As discussed above, it is clear that $|V_p|=N/C$. As an extreme case suppose there are only a few classes in the dataset, i.e., $C$ is small enough. In this case, it can be derived that $|V_p|=\frac{N}{C}\approx N$, the scale of potential nodes comes larger and unbearable. Consider that we only have a budget $\Delta$ for a target node, which means that we can add $\Delta$ edges at most. To further improve the efficiency, we will eventually leave $\Delta$ potential nodes $\hat{V}_p\subseteq V_p$, where the gradients of potential edges between $t$ and them are $\Delta$-largest, i.e.
\begin{equation}
  \label{nodereduction}
  \begin{aligned}
    \hat{V}_p & =\{u\,|\, \nabla _{G_{t,u}^{(sub)}} \, \mbox{is} \,\Delta\, \mbox{-largest},\  u \in V_p\}, \\
    \hat{E}_p & =\mathcal{A} \times \hat{V}_p,
  \end{aligned}
\end{equation}
where $\nabla _{G_{t,u}^{(sub)}}$ is the gradient of loss $\Tilde{\mathcal{L}}_t^{(sub)}$ w.r.t. the edge $(t,u)$.

Let $V^{(sub)}=V_s \cup \hat{V}_p$,  $E^{(sub)}=E_s \cup \hat{E}_p$, the scale of the subgraph  $G^{(sub)}$ comes smaller. We only need to compute $\nabla _{G_{u,v}^{(sub)}} $ for each (non-)edge $(u,v) \in E^{(sub)}$. Either the input graph is sparse or not, it's obvious that $|E^{(sub)}|\ll |E|$. Besides, computing $Z^{(sub)}_t$ using $A^{(sub)}$ can be done in constant time and it further improves the efficiency.

Figure \ref{SGA} shows the simplified gradient-based attack via subgraph, where we not only extract the $k$-hop subgraph centered at the target node but also add some potential nodes to extend the subgraph. By doing so, we can enlarge the possible perturbation set and compute the gradients of potential edges in this extended subgraph, hence our method is available to add adversarial edges as well.

\begin{figure}
  \centering
  \subfigure[Citeseer] {\includegraphics[width=0.45\linewidth]{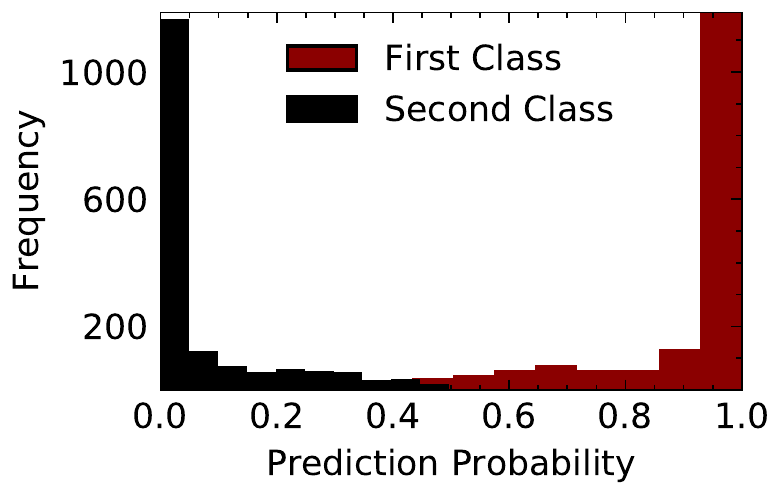}}\hspace{1mm}
  \subfigure[Cora] {\includegraphics[width=0.45\linewidth]{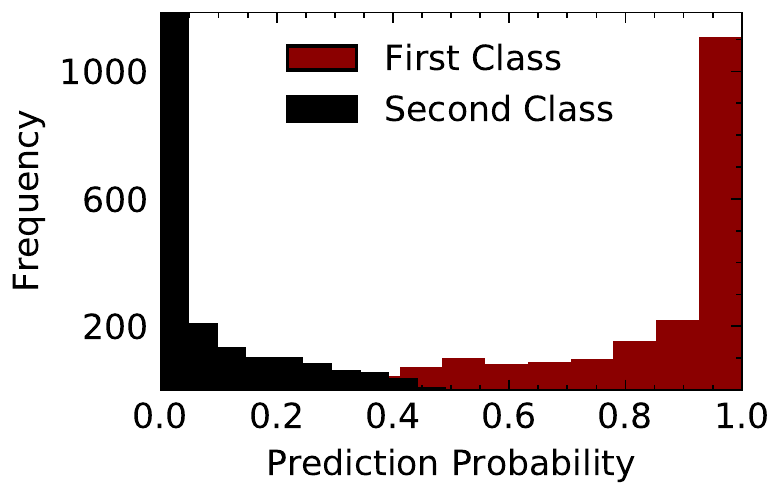}}\hspace{1mm}
  \subfigure[Citeseer (calibrated)] {\includegraphics[width=0.45\linewidth]{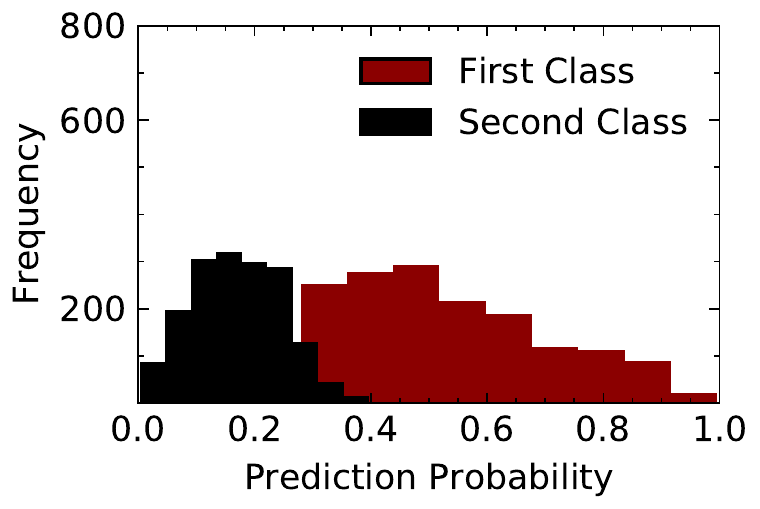}}\hspace{1mm}
  \subfigure[Cora (calibrated)] {\includegraphics[width=0.45\linewidth]{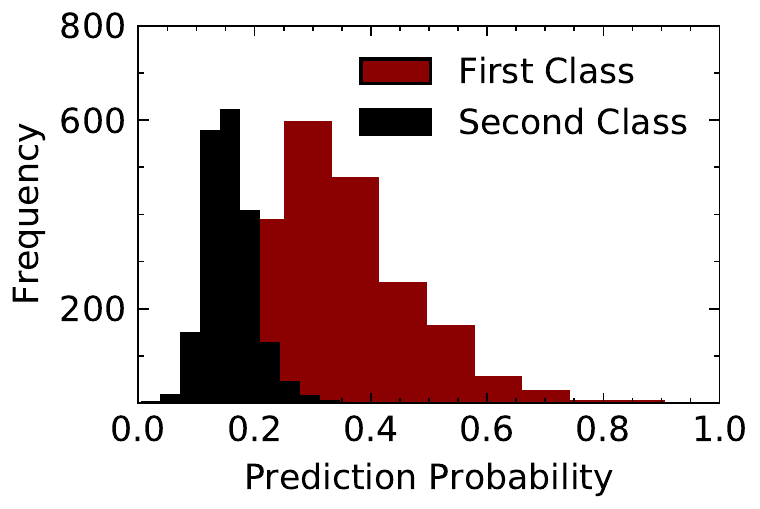}}\hspace{1mm}
  \caption{The prediction probability of the first and next most probable (second) class of SGC on Citeseer and Cora datasets before and after calibration.}
  \label{prob_distribution}
\end{figure}

\subsection{Gradient Computation with Calibration}
\label{sec:gradient}
Gradient-based algorithm is an efficient and effective method to attack. Similar to GradArgmax, the gradients of the targeted loss w.r.t the subgraph will be computed as follows:
\begin{equation}
  \begin{aligned}
    \label{gradients}
     & \nabla _{G^{(sub)}} = \nabla _{A^{(sub)}} \Tilde{\mathcal{L}}_t^{(sub)}\,,
  \end{aligned}
\end{equation}
% where $\Tilde{\mathcal{L}}_t^{(sub)}=\max_{c^\prime_t \neq c_t} \ln{Z_{t,c^\prime_t}^{(sub)}} - \ln{Z_{t,c_t}^{(sub)}}$ denotes the targeted loss computed using subgraph and it is derived from Eq.(\ref{sgaloss}), $Z_t^{(sub)} = f(A^{(sub)},X) \in \mathbb{R}^{1\times C}$ denotes the prediction probability vector of target node $t$ using subgraph. Note that, to ensure $\hat{A}^{(sub)}_{u,v}= \hat{A}_{u,v}, \forall u, v \in V_s$, we must add self-loop for each node $u \in V_s$, and normalize $\Tilde{A}^{(sub)}$ using $\Tilde{D}$ the same as $\Tilde{A}$. Following this, apparently $Z_t^{(sub)}=Z_t$ according to Property 1. As described in Eq.(\ref{gradients}), the partial derivative of $\Tilde{\mathcal{L}_t}$ w.r.t the adjacency matrix ${A}^{(sub)}$ is computed for each edge $e=(u,v)\in E_s$.

Note that, to ensure $\hat{A}^{(sub)}_{u,v}= \hat{A}_{u,v}, \forall u, v \in V_s$, we must add self-loop for each node $u \in V^{(sub)}$, and normalize $\Tilde{A}^{(sub)}$ using $\Tilde{D}$ the same as $\Tilde{A}$. Following this, apparently $Z_t^{(sub)}=Z_t$ according to Property 1. As described in Eq.(\ref{gradients}), the partial derivative of $\Tilde{\mathcal{L}_t}^{(sub)}$ w.r.t the adjacency matrix ${A}^{(sub)}$ is computed for each (non-)edge $e=(u,v)\in E^{(sub)}$.

However, as stated by Guo et al. \cite{Guo2017OnCO}, most of the modern neural networks are poorly calibrated, and the probability associated with the predicted class label doesn't reflect its ground-truth correctness likelihood. On the contrary, \emph{a neural network appears to be overconfident on the predictions}. As shown in Figure \ref{prob_distribution}(a) and Figure \ref{prob_distribution}(b), the model is so confident that it gives very high probability for the predicted class ($Z_{t, c_t} \rightarrow 1$), but for other classes even the next most probable one, it gives an extremely low probability ($Z_{t, c^\prime_t} \rightarrow 0$). As a result, the first term of targeted loss $\Tilde{\mathcal{L}}_t^{(sub)}$ rounds to minus infinity while the second term rounds to zero. This results in the gradients vanishing and thus the attack performance is negatively affected. To overcome the gradient vanishing problem, a scale factor $\epsilon>1$ is introduced to calibrate the output:
% , and the targeted loss is redefined as follows:

\begin{equation}
  \begin{aligned}
    \label{targetedloss}
     & Z_t^{(sub)} = f_\theta (A^{(sub)},X, \epsilon)=\mathrm{softmax}\,(\frac{\hat{A}^{(sub)k}XW}{\epsilon}).
    %  & \Tilde{\mathcal{L}}_t^{(sub)}= \ln{Z_{t,c^\prime_t}^{(sub)}} - \ln{Z_{t,c_t}^{(sub)}},\quad t\in V
  \end{aligned}
\end{equation}
By introducing a scale factor $\epsilon$, the output of the model will no longer produce the extreme value 0 or 1 (See Figure \ref{prob_distribution}(c) and Figure \ref{prob_distribution}(d)), thereby avoid the gradients vanishing.

\subsection{Iterative Gradient-based Attack}
\label{sec:update}
To better capture the actual change of the surrogate loss, we will sequentially compute the gradients after adding or removing an edge \cite{zugner2019adversarial}. Let $Z_t^{(i)}$ and $ Z_t^{(sub_i)}$ denote the prediction of target node $t$ at the $i_{th}$ time using the whole graph and subgraph, respectively. ${A}^{(sub_i)}$, $V^{(sub_i)}$, $E^{(sub_i)}$ are defined in the same way. Specifically, ${A}^{(sub_0)}={A}^{(sub)}$, $V^{(sub_0)}=V^{(sub)}$, $E^{(sub_0)}=E^{(sub)}$.

\begin{figure}[t]
  \centering
  \includegraphics[width=\linewidth]{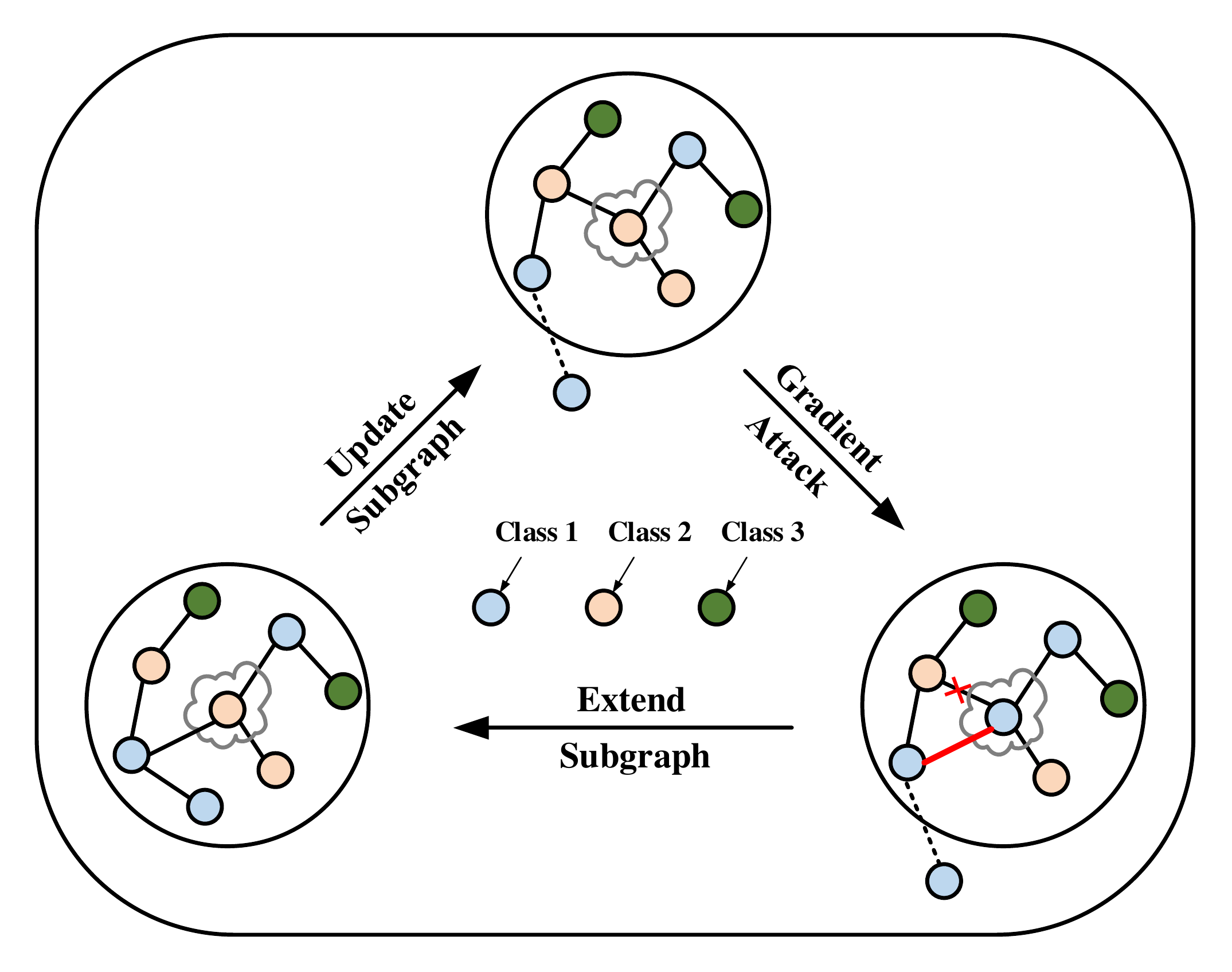}
  \caption{Iterative gradient-based attack via subgraph expansion ($k=2$). The largest circle denotes the radius ($k$) of neighborhood centered at target node. If a node becomes closer to the target node within $k$-distance, the edges connected to it should be considered as well.}
  \label{iterative_SGA}
\end{figure}

At each time $i$+1, we flip one edge $e=(u,v) \in E^{(sub_{i})}$ with the \emph{largest}
magnitude of gradient that fulfills the constraints \cite{chen2020link}:
\begin{equation}
  \label{constraints}
  \begin{cases}
    \mbox{Connect}\ (u,v),    & \mbox{if }\  \nabla _{G^{(sub_i)}}>0\ \mbox{and}\ A^{(sub_i)}_{u,v}=0 \\
    \mbox{Disconnect}\ (u,v), & \mbox{if }\  \nabla _{G^{(sub_i)}}<0\ \mbox{and}\ A^{(sub_i)}_{u,v}=1
  \end{cases}
\end{equation}
To conclude, we define the structure score as follows to simplify Eq.(\ref{constraints}):
\begin{equation}
  \label{structure_score}
  S = \nabla _{G^{(sub_i)}} \odot (-2A^{(sub_i)}+1)\,,
\end{equation}
where $\odot$ denotes Hadamard product. The adversarial edge $(u,v)$ is selected based on the \emph{largest} structure score. After an edge is selected, we update the subgraph $G^{(sub_{i+1})}$ as follows:
\begin{equation}
  \begin{aligned}
    V^{(sub_{i+1})} & =V^{(sub_{i})},          \\
    E^{(sub_{i+1})} & =E^{(sub_{i})} - \{e\},  \\
    A^{(sub_{i+1})} & =A^{(sub_{i})} \oplus e,
  \end{aligned}
\end{equation}
where $\oplus$ denotes the ``exclusive or'' operation,  for $e=(u,v)$, if $A^{(sub_{i})}_{u,v}=0$ then $A^{(sub_{i+1})}_{u,v}=1$ and vice versa. However, it will results in $Z^{(sub_i)}_t \neq Z^{(i)}_t, \forall i > 0$. We explain it with Figure  \ref{iterative_SGA},  when a potential edge is connected, if a node becomes closer to the target node within $k$-distance, the edges adjacent to it should be considered as well, so as to \emph{preserve the $k$-order  message aggregation of graph convolution}. Therefore, if an edge $e=(u,v)\in E^{(sub_i)}$ is connected (assume that $v$ is a potential node), we must expand the subgraph as follows:
\begin{gather}
  \label{update_rules}
  \begin{aligned}
    V^{(sub_{i+1})} & =V^{(sub_{i})} \cup \{{v^\prime}\,|\,{v^\prime} \in \mathcal{N}(v) \ \mbox{if}\ \mathcal{D}(t,v^\prime) \leq k\}, \\
    E^{(sub_{i+1})} & =E^{(sub_{i})} \cup (E^{(exp)} - \{e\}),                                                                          \\
    A^{(sub_{i+1})} & =A^{(sub_{i})} \oplus  (E^{(exp)} \cup \{e\}),
  \end{aligned}
\end{gather}
where $E^{(exp)}=\{(v,{v^\prime})\,|\,{v^\prime} \in \mathcal{N}(v)\ \mbox{if}\ \mathcal{D}(t,v^\prime) \leq k\}$ is the expended edges set where nodes become closer within $k$-distance.

%%算法伪代码
\begin{algorithm}[t]
  \caption{Simplified gradient-based attack (SGA)}
  \label{algorithm1}
  \begin{algorithmic}[1] %这个1 表示每一行都显示数字
    \REQUIRE ~~\\ %算法的输入参数：Input
    Graph $G=(A,X)$, attack budget $\Delta$, target node $t$, ground-truth label $c_t$, labeled nodes $V_L$, hyper-parameter $k$ for surrogate model SGC;
    \ENSURE ~~\\ %算法的输出：Output
    Perturbed graph $G_{t}^\prime$ w.r.t target node $t$;
    \STATE $\theta \leftarrow$ train the surrogate model on labeled nodes $V_L$;
    \STATE $c^\prime_t =\arg \max_{c^\prime\neq c_t} Z_{t,c^\prime} \leftarrow$ predict the next most probable class of $t$ ;
    \STATE Extract the $k$-hop subgraph centered at $t$ ;
    \STATE  Initialize the subgraph $G^{(sub)}=(A^{(sub)},X)$ via Eq.(\ref{nodereduction});
    \STATE $A^{(sub_0)}$, $E^{(sub_0)}$, $V^{(sub_0)}$ $\leftarrow$ $A^{(sub)}$, $E^{(sub)}$, $V^{(sub)}$;
    \FOR {$i=0$ to $\Delta - 1$}
    {
    \STATE $Z_t^{(sub_i)}=f_\theta (A^{(sub_i)},X, \epsilon)$;
    \STATE Compute $\nabla _{G^{(sub_i)}}$ with Eq.(\ref{gradients});
    \STATE Compute structure score $S$ with Eq.(\ref{structure_score});
    \STATE Select $e=(u,v)$ with largest structure score $S_{u,v}$;
    \STATE Update $A^{(sub_{i+1})}$, $E^{(sub_{i+1})}$, $V^{(sub_{i+1})}$ with Eq.(\ref{update_rules});
    }
    \ENDFOR
    \label{ code:fram:extract }%对此行的标记，方便在文中引用算法的某个步骤
    \RETURN $G_{t}^\prime$; %算法的返回值
  \end{algorithmic}
\end{algorithm}

Since the scale of potential nodes is small enough, the subgraph expansion does not negatively affect the efficiency$\footnote{In contrast, if an edge is disconnected from the subgraph, the message aggregation will be blocked automatically for those nodes beyond $k$-distance.}$. The details of iterative SGA is summarized in Algorithm \ref{algorithm1}.

%Therefore, those edges and nodes can be either removed or not from the subgraph since they do not make any difference. In this paper, it is omitted due to space constraints.

%% 时空复杂度
\subsection{Analysis of Time and Space Complexity}
To better describe our method, we compare the complexity of time and space with other state-of-the-art methods: \emph{Nettack} and \emph{GradArgmax}.

\subsubsection{Nettack}
Due to store the whole graph (as a sparse matrix), the space complexity of Nettack comes to $\mathcal{O}(|E|)$. Besides, Nettack computes the misclassification loss sequentially w.r.t the perturbed graph for each edge in the candidate set, the time complexity is up to $\mathcal{O}(\Delta \cdot |\mathcal{A}| \cdot N\cdot d^2)$ without considering feature attack, where $d$ denotes the average degree of nodes in the graph.

\subsubsection{GradArgmax}
While storing the whole graph with a sparse matrix, the space complexity of GradArgmax will be $\mathcal{O}(|E|)$. If a dense instantiation of the adjacency matrix is used, the space complexity comes to $\mathcal{O}(N^2)$. It's worth mentioning that, for the adjacency matrix in sparse form, the algorithm is able to delete edges only, while both adding and deleting edges are available for the dense one. Besides, GradArgmax only computes gradients once, so the major time complexity depends on the computation of gradients and it is approximate $\mathcal{O}(\Delta + |E|d^2)$ or $\mathcal{O}(\Delta + N^2d^2)$ for sparse or dense matrix$\footnote{Refer to \cite{goyal2018graph}, the time complexity of GCN is $\mathcal{O}(|E|d^2)$, and the computation of gradients can be accelerated by parallel computing, which is also suitable for SGA.}$.

\subsubsection{SGA}
Our method only stores the subgraph with a sparse matrix, thus the space complexity is only $\mathcal{O}(|E^{(sub)}|)$, where $E^{(sub)}$ only consists of edges of $k$-hop subgraph and a few potential edges, which comes to $|E^{(sub)}|=d^k+\Delta \cdot |\mathcal{A}|$, apparently $|E^{(sub)}| \ll |E|$ for a sparse graph. Moreover, the time complexity depends on the message aggregation between the $k$-hop neighbors of the target node in $A^{(sub)}$, thereby the time complexity leads to $\mathcal{O}(\Delta \cdot |E^{(sub)}|\cdot d^k)$. Given $|E^{(sub)}|=d^k+\Delta \cdot |\mathcal{A}|$, it can be simplified as $\mathcal{O}(\Delta \cdot (d^k+\Delta \cdot |\mathcal{A}|)\cdot d^k)$. The complexity of the three methods are summarized in Table \ref{complexity}, where the usage of memory refers to the usage of graph (subgraph). It is clear that SGA theoretically achieves both time and space efficiency compared to Nettack and GradArgmax.

% \begin{figure}[t]
%   \centering
%   \includegraphics[width=\linewidth]{fig5.pdf}
%   \caption{Typical Example of Assortative or Disassortative networks. The fuzzy edges indicate that nodes can have any number of edges of this type. (Image Credit: Foster et al. \protect\cite{foster2010edge}) }
%   \label{figure4}
% \end{figure}

\begin{table}[t]
  \centering
  \setlength{\belowcaptionskip}{10.0pt}
  \caption{Time and Space Complexity of Adversarial Attack Algorithms.}
  \label{complexity}
  \begin{tabular}{c|c|c}
    % \toprule
    \hline
    Methods          &
    Space complexity &
    Time complexity                                                                                                                          \\
    % \midrule
    \hline
    \hline
    Nettack          & $\mathcal{O}(|E|)$                            & $\mathcal{O}(\Delta \cdot |\mathcal{A}| \cdot N\cdot d^2)$            \\
    GradArgmax       & $\mathcal{O}(|E|)$ or $\mathcal{O}(N^2)$      & $\mathcal{O}(\Delta + |E|d^2)$ or $\mathcal{O}(\Delta + N^2d^2)$      \\
    SGA              & $\mathcal{O}(d^k+\Delta \cdot |\mathcal{A}|)$ & $\mathcal{O}(\Delta \cdot (d^k+\Delta \cdot |\mathcal{A}|)\cdot d^k)$ \\
    % \bottomrule
    \hline
  \end{tabular}
\end{table}

\section{UNNOTICEABLE ADVERSARIAL ATTACK ON GRAPH DATA}
Specifically, in the scenario of adversarial attack, attackers would attempt to perform perturbations by modifying the input data while guaranteeing \emph{unnoticeability} to avoid the detection. Unlike the continuous data that lies in \emph{Euclidean Space}, attackers can evaluate the attack impacts by measuring the changes before and after attack via $\ell_2$ norm or $\ell_{\infty}$ norm \cite{madry2017towards}. However, it is difficult to quantify the attack impacts on graph data that lies in such a \emph{non-Euclidean Space}.

An important property of a graph is the degree distribution, which often appears as a power-law distribution \cite{bessi2015two}, i.e., $p(x) \propto x^{-q}$, where $q$ is the scaling parameter. It is easy to tell if two graphs have similar degree distributions after $q$ is given. However, there is no solution for estimating $q$ exactly yet. To this end,  Z\"{u}gner et al. \cite{zugner2018adversarial} derive an efficient way to check for violations of the degree distribution after attacks. Specifically, they estimate $q$ for a clean graph and $q^\prime$ for a perturbed graph to ensure that the attack is unnoticeable if the following equation is fulfilled:
\begin{gather*}
  \Lambda \,(q, q^\prime; G, G^\prime)<\tau \approx 0.004\,,
\end{gather*}
where $\Lambda$ is a discriminate function defined in \cite{zugner2018adversarial}, $G$ denotes the original graph and $G^\prime$ the perturbed one.

In this section, we refer to another solution and apply it to measure the attack impacts, \emph{degree assortativity coefficient} \cite{newman2003mixing,foster2010edge}, it measures the tendency of nodes connected to each other. To clarify this, we first define the degree mixing matrix $M$, where $M_{i,j}$ denotes the tendency of nodes with degree $i$ connected to nodes with degree $j$, the value can be counts, joint probability, or occurrences of node degree. Let $\alpha, \beta \in \{in,out\}$ denote the type of in-degree and out-degree, respectively. We first ensure that $M$ fulfills the following rules:
\begin{equation}
  \sum_{ij} M_{i,j}=1,\quad \sum_j M_{i,j}=\alpha_i,\quad \sum_i M_{i,j}=\beta_j\,,
\end{equation}
Particularly, in an undirected graph, $\alpha_i=\beta_j,\, \forall i=j$. Then we define the degree assortativity coefficient by using Pearson Correlation \cite{goh2007human}:
\begin{equation}
  r(\alpha,\beta)=\frac{\sum_{ij}ij(M_{i,j}-\alpha_i \beta_j)}{\sigma_\alpha\sigma_\beta},
\end{equation}
where $\sigma_\alpha$ and $\sigma_\beta$ are the standard deviations of the distributions $\alpha$ and $\beta$, respectively. The value of $r$ lies in the range of $-1\leq r \leq 1$, and $r=-1$ for disassortativity while $r=1$ for assortativity.

% Figure \ref{figure4} shows the four degree-degree correlations in a directed network, which reflects the correlation between nodes in the network as well.

Degree assortativity coefficient $r$ is much easier to compute than degree distribution coefficient $q$, and it better reflects the graph structure. Based on this, we propose \textbf{Degree Assortativity Change (DAC)}, which is defined as
\begin{equation}
  \mathrm{DAC}=\frac{\mathbb{E}_r(\,|r_{G}-r_{G_{t_i}'}|\,)}{r_{G}},\quad \forall t_i
\end{equation}
where $G_{t_i}'$ denotes the perturbed graph w.r.t target node $t_i$. Attackers will conduct attack for each target node $t_i$ by performing perturbations on the original graph $G$. DAC measures the average impacts on attacking a group of target nodes $\{t_i\}$, and the smaller the DAC is, the less noticeable the attack will be.

\section{EXPERIMENTS}
In this section, we conduct extensive experiments aiming to answer the following research questions:
\begin{itemize}
  \item[\textbf{RQ1}] How much can SGA improve in terms of time and space efficiency and how does it work for the proposed metric DAC?
  \item[\textbf{RQ2}] Can SGA achieve a competitive attack performance while using only a certain part of nodes to attack?
  \item[\textbf{RQ3}] Can SGA scale to larger datasets and maintain considerable attack performance?
\end{itemize}
In what follows, we first detail the experimental settings, followed by answering the above three research questions.

\subsection{Experimental Settings}
\textbf{\emph{Datasets}}. We evaluate the performance of our method on four well-known datasets: Citeseer, Cora, Pubmed \cite{sen2008collective} and Reddit \cite{hamilton2017inductive}. The first three are commonly used citation networks, where nodes are documents and edges among them are citation relations. The last one is a large scale dataset of online discussion forums where user posts and comments on content in different topical communities. We follow the setting of Nettack \cite{zugner2018adversarial} and only consider the largest connected component of the graph for each dataset. Besides, we randomly select $20\%$ of nodes to constitute the training set (half of them are treated as the validation set) and treat the remaining as the testing set. Table \ref{dataset} is an overview of the datasets.

\textbf{\emph{Target Model}}. We consider poisoning attacks, i.e., models are retrained on the perturbed graph until the convergence after attacks \cite{chen2020survey}, which is more challenging for attackers but reflects real-world scenarios better. To evaluate the transferability of our method, we conduct attack on several commonly used graph neural networks:  GCN \cite{kipf2016semi}, SGC \cite{wu2019simplifying}, GAT \cite{velickovic2018graph}, GraphSAGE \cite{hamilton2017inductive}, ClusterGCN \cite{chiang2019cluster}. For each target model, our method is compared with other adversarial attack methods.
\begin{itemize}
  \item \textbf{GCN} \cite{kipf2016semi}. GCN is one of the most representative graph neural networks that learn hidden representations by encoding both local graph structure and node features.
  \item \textbf{SGC} \cite{wu2019simplifying}. SGC is a linear variant of GCN that has been proposed recently, which achieves competitive results and even significantly improves the training efficiency.
  \item \textbf{GAT} \cite{velickovic2018graph}. GAT enhances GCN by leveraging a masked self-attention mechanism to specify different weights to different neighbor nodes.
  \item \textbf{GraphSAGE} \cite{hamilton2017inductive}. GraphSAGE is a general inductive framework, which uniformly samples a set of neighbors with a fixed size, instead of using a full-neighborhood set during training.
  \item \textbf{ClusterGCN} \cite{chiang2019cluster}. This is the state-of-the-art mini-batch GCN framework. It samples $n$ subgraphs whose nodes have high correlations under a graph clustering algorithm and restricts the message aggregation within these subgraphs.
\end{itemize}

\begin{table}[t]
  \centering
  \setlength{\belowcaptionskip}{10.0pt}
  \caption{Dataset statistics. Only consider the largest connected component of the graph for each dataset.}
  % \fontsize{8}{8}\selectfont
  \begin{tabular}{c|c|c|c|c}
    % \toprule
    \hline
    Statistics     & Citeseer & Cora    & Pubmed  & Reddit     \\
    % \midrule
    \hline
    \hline
    \#Nodes        & 2,110    & 2,485   & 19,717  & 232,965    \\
    \#Edges        & 3,668    & 5,069   & 44,324  & 11,723,402 \\
    Density        & 0.082\%  & 0.082\% & 0.011\% & 0.022\%    \\
    Average Degree & 3.50     & 4.08    & 4.50    & 99.65      \\
    % \bottomrule
    \hline
  \end{tabular}
  \label{dataset}
\end{table}

\begin{table*}[t]
  \centering
  \setlength{\belowcaptionskip}{10.0pt}
  %   \setlength{\belowcaptionskip}{5.0pt}
  %   \fontsize{7}{7}\selectfont
  \caption{Average time (s), memory usage, and DAC to generate adversarial examples using both attack strategies. The statistics of random algorithms RA and DICE are omitted. OOM stands for a method that could not be used to attack due to the limitations of GPU RAM.}
  \label{time_mem_DAC}
  \resizebox{\textwidth}{!}{
    \begin{tabular}{c|ccc|ccc|ccc|ccc}
      % \toprule
      \hline
      Methods          & \multicolumn{3}{c|}{Citeseer } & \multicolumn{3}{c|}{Cora} & \multicolumn{3}{c}{Pubmed} & \multicolumn{3}{c}{Reddit}\cr
      % \cmidrule(lr){2-7}
      \hline
      \hline
      Direct Attack    & Time                           & Memory                    & DAC                        & Time                          & Memory & DAC    & Time    & Memory & DAC    & Time   & Memory & DAC \cr
      % \midrule
      \hline
      GradArgmax       & 0.188                          & 17MB                      & 7.9E-2                     & 0.297                         & 23MB   & 2.9E-3 & 13.856  & 1483MB & 4.2E-3 & N/A    & OOM    & -\cr
      Nettack          & 0.722                          & 66KB                      & 3.2E-2                     & 1.151                         & 66KB   & 1.4E-3 & 39.526  & 769KB  & 1.3E-3 & N/A    & 180MB  & -\cr
      % \midrule
      \hline
      SGA              & 0.006                          & 2KB                       & 3.4E-2                     & 0.011                         & 2KB    & 1.7E-3 & 0.020   & 4KB    & 1.8E-3 & 12.236 & 14MB   & 2.5E-6\cr

      % \bottomrule
      % \toprule
      \hline
      \hline

      Influence Attack & Time                           & Memory                    & DAC                        & Time                          & Memory & DAC    & Time    & Memory & DAC    & Time   & Memory & DAC\cr
      % \midrule
      \hline
      GradArgmax       & 0.239                          & 17MB                      & 1.7E-2                     & 0.321                         & 23MB   & 1.1E-3 & 14.412  & 1483MB & 9.2E-4 & N/A    & OOM    & -\cr
      Nettack          & 2.104                          & 89KB                      & 1.6E-2                     & 3.969                         & 89KB   & 9.4E-4 & 120.821 & 769KB  & 8.1E-4 & N/A    & 180MB  & -\cr
      \hline
      SGA              & 0.008                          & 5KB                       & 1.7E-2                     & 0.012                         & 6KB    & 1.0E-3 & 0.021   & 7KB    & 8.7E-4 & 14.571 & 14MB   & 1.9E-6\cr

      % \bottomrule    
      \hline
    \end{tabular}
  }
\end{table*}

\textbf{\emph{Evaluation Protocols}}. For the task of node classification, we aim to perform perturbations and further cause misclassification of target models. To this end, we evaluate the classification accuracy and classification margin (CM). Given a target node $t$, CM is defined as the probability margin between ground-truth label $c_t$ and the next most probable class label $c^\prime_t$, which lies in the range of $[-1,1]$. The smaller CM means better attack performance. In particular, a successful attack is often with CM less than zero.

\textbf{\emph{Baselines}}.
We compare our method with four other baseline methods and conduct attacks using two strategies respectively (i.e., direct attack and influence attack). For a fair comparison, all methods will use the same surrogate model SGC (if necessary) and share the same weights $\theta$. Follow the setting of Nettack \cite{zugner2018adversarial}, the attack budget $\Delta$ is set to the degrees of target node $t$.
\begin{itemize}
  \item \textbf{Random Attack (RA)}. RA randomly adds or removes edges between attacker nodes and other nodes in the graph with probability $p_1$. This is the simplest way to conduct attacks.
  \item \textbf{DICE} \cite{cai2005mining}. DICE is originally a heuristic algorithm for disguising communities. In our experiment, DICE randomly decides whether to connect or disconnect an edge with probability $p_2$ between attacker nodes and other nodes in the graph. Besides, there is a constraint that only nodes belonging to the same class/different classes will be disconnected/connected.
  \item \textbf{GradArgmax} \cite{dai2018adversarial}. Since the attack budget $\Delta$ is defined as the degrees of target node $t$, the original GradArgmax will remove all edges connected with $t$ for direct attack, which is unreasonable and unfair to compare. Therefore, we use the variant of GradArgmax, which requires a dense instantiation of the adjacency matrix and computes the gradients of all $N^2$ edges.
  \item \textbf{Nettack} \cite{zugner2018adversarial}. Nettack is the strongest baseline that can modify the graph structure and node features. As we focus on the structure attack, we restrict Nettack to modify the graph structure with budget $\Delta$ only.
\end{itemize}

\label{parm}
\textbf{\emph{Parameter Settings}}.
The hyper-parameters of target models are fine-tuned in the clean graph for each dataset. Particularly, for SGA, the radius $k=2$ and the scale factor $\epsilon=5.0$ for each dataset, and the number of added potential nodes is set to $\Delta$ when generating adversarial examples. For RA and DICE, $p_1$ and $p_2$ are both fixed at $0.5$. All models are implemented in Tensorflow$\footnote{\url{https://www.tensorflow.org/}}$,  running on a NVIDIA RTX 2080Ti GPU.

\textbf{\emph{Attack Settings}}. Our experiments involve two parts: (i) \emph{Generating adversarial examples}. For each dataset, we first randomly select 1,000 nodes from the testing set as target nodes and generate adversarial examples for them separately using different attack methods. (ii) \emph{Conduct attack through adversarial examples}. After adversarial examples are generated, we conduct poisoning attacks on the surrogate model SGC, and further transfer it to other commonly used graph neural networks.

\begin{figure}
  \centering
  \includegraphics[width=\linewidth]{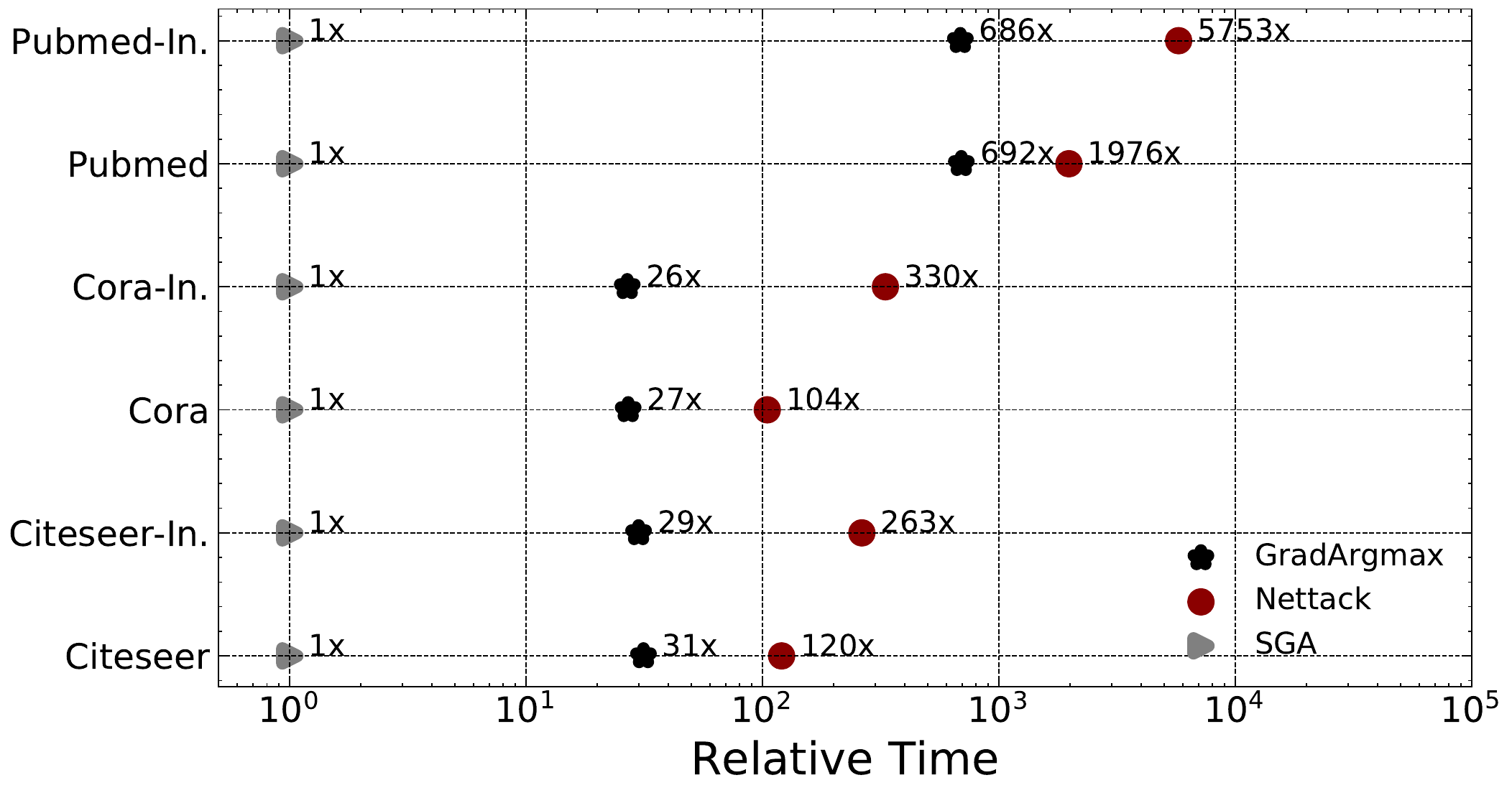}
  \caption{Performance over running time on three datasets for direct and influence attack.}
  \label{fig_times}
\end{figure}

\begin{table*}[t]
  \setlength{\belowcaptionskip}{10.0pt}
  \centering
  \caption{Accuracy(\%) of direct attack and influence attack on two small scale datasets. Here the best results are \textbf{boldfaced}.}
  \label{attack_small}
  \resizebox{\textwidth}{!}{
    \begin{tabular}{c|ccccc|ccccc}
      % \toprule
      \hline
      Methods          & \multicolumn{5}{c|}{Citeseer} & \multicolumn{5}{c}{Cora}                                                                                                                                     \\
      \hline
      \hline
      Direct Attack    & SGC                           & GCN                      & GAT           & GraphSAGE     & Cluster-GCN   & SGC           & GCN           & GAT           & GraphSAGE     & Cluster-GCN       \\
      % \midrule
      \hline
      Clean            & 71.8                          & 71.6                     & 72.1          & 71.0          & 71.4          & 83.1          & 83.5          & 84.2          & 82.0          & 83.2              \\
      RA               & 62.0                          & 63.6                     & 59.0          & 61.9          & 62.8          & 67.5          & 68.8          & 69.7          & 68.8          & 68.1              \\
      DICE             & 55.0                          & 57.8                     & 54.6          & 56.6          & 57.1          & 58.4          & 60.2          & 61.8          & 59.7          & 59.5              \\
      GradArgmax       & 12.2                          & 13.2                     & 21.5          & 32.4          & 34.8          & 25.2          & 32.6          & 34.5          & 30.9          & 45.6              \\
      Nettack          & 3.7                           & 6.0                      & 20.5          & \textbf{27.9} & 30.9          & \textbf{1.0}  & 2.4           & 17.6          & 27.2          & 17.8              \\

      % \midrule
      \hline
      SGA              & \textbf{1.8}                  & \textbf{3.8}             & \textbf{20.3} & 30.2          & \textbf{19.9} & 1.5           & \textbf{2.1}  & \textbf{15.9} & \textbf{25.8} & \textbf{17.6}     \\
      % \bottomrule
      % \toprule
      \hline
      \hline
      Influence Attack & SGC                           & GCN                      & GAT           & GraphSAGE     & Cluster-GCN   & SGC           & GCN           & GAT           & GraphSAGE     & Cluster-GCN   \cr
      % \midrule
      \hline
      RA               & 71.2                          & 71.1                     & 71.4          & 68.8          & 70.8          & 80.9          & 83.0          & 86.0          & 81.4          & 83.0              \\
      DICE             & 69.4                          & 71.1                     & 70.2          & 67.5          & 68.4          & 80.5          & 82.4          & 85.4          & 79.8          & 82.7              \\
      GradArgmax       & 47.8                          & 48.1                     & 51.2          & 46.7          & 55.4          & 65.1          & 70.8          & 73.8          & 74.7          & 72.9              \\
      Nettack          & \textbf{31.4}                 & 39.2                     & 49.3          & 41.9          & 49.8          & 48.4          & 56.4          & 63.2          & 63.5          & 60.0              \\

      % \midrule
      \hline
      SGA              & 33.1                          & \textbf{38.5}            & \textbf{45.2} & \textbf{40.2} & \textbf{48.1} & \textbf{46.2} & \textbf{56.2} & \textbf{62.6} & \textbf{62.8} & \textbf{57.1}     \\
      % \bottomrule    
      \hline
    \end{tabular}
  }
\end{table*}

\begin{figure*}
  \centering
  \subfigure[SGC] {\includegraphics[width=0.185\linewidth]{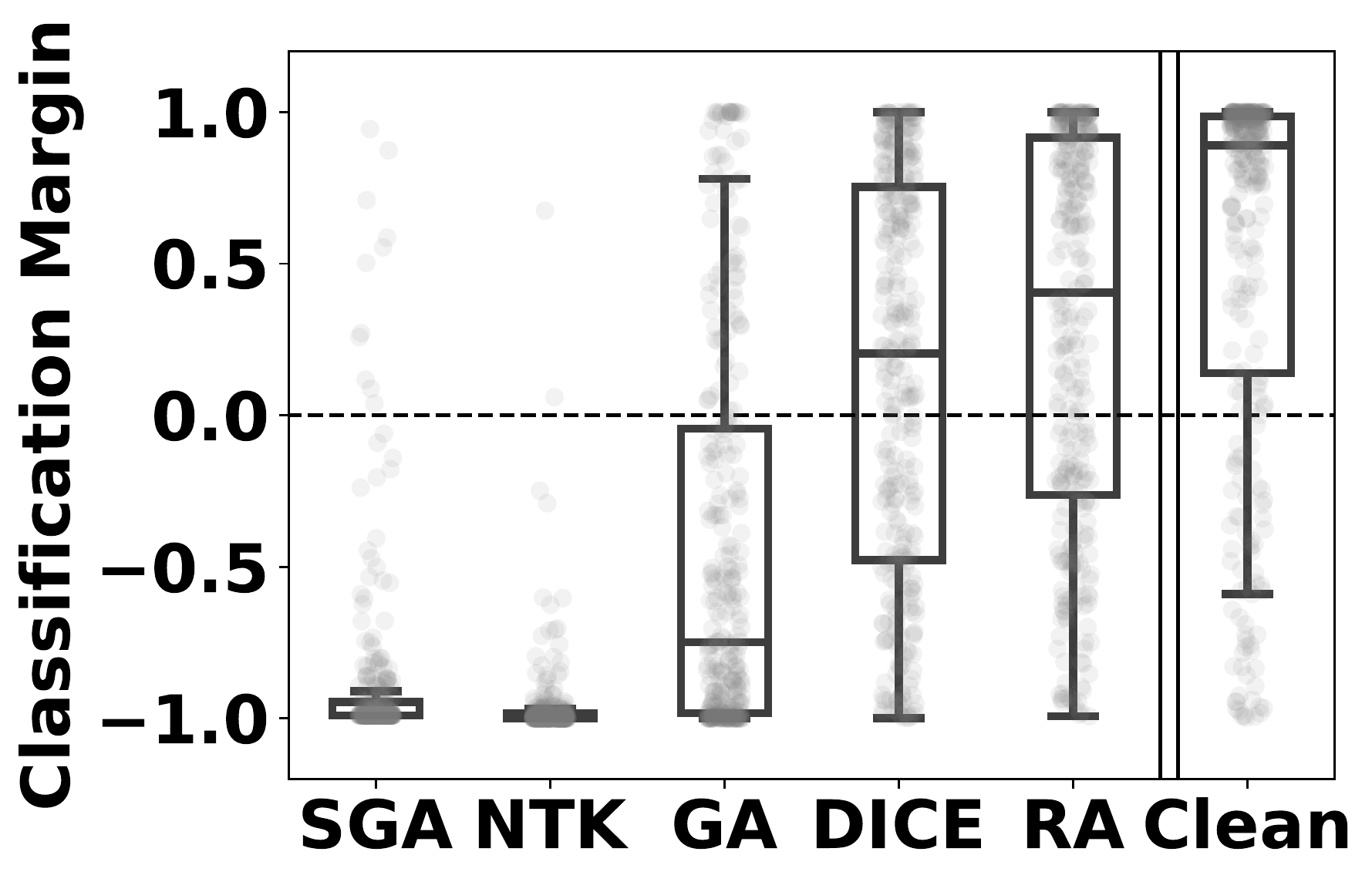}}\hspace{1mm}
  \subfigure[GCN] {\includegraphics[width=0.185\linewidth]{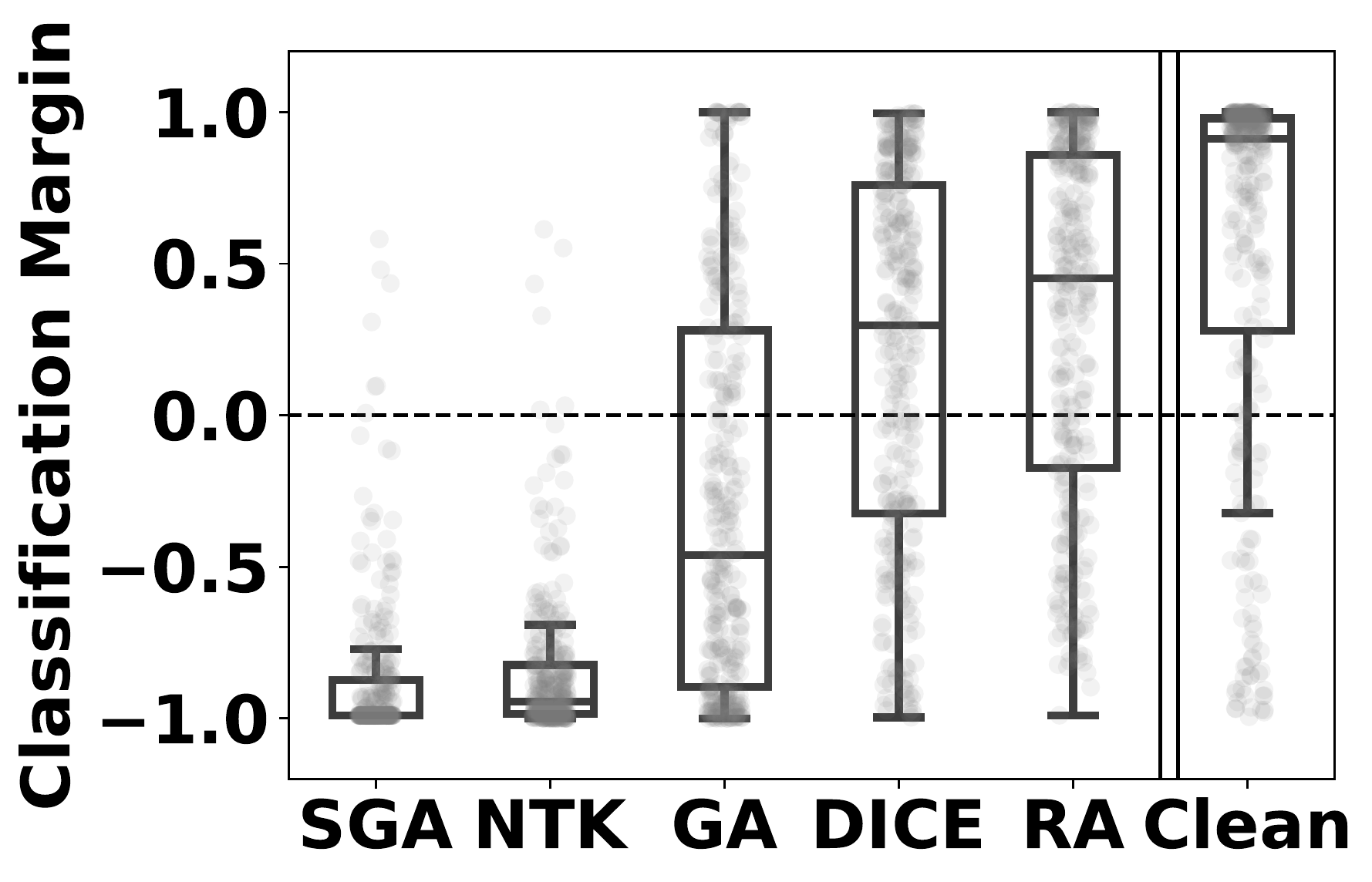}}\hspace{1mm}
  \subfigure[GAT] {\includegraphics[width=0.185\linewidth]{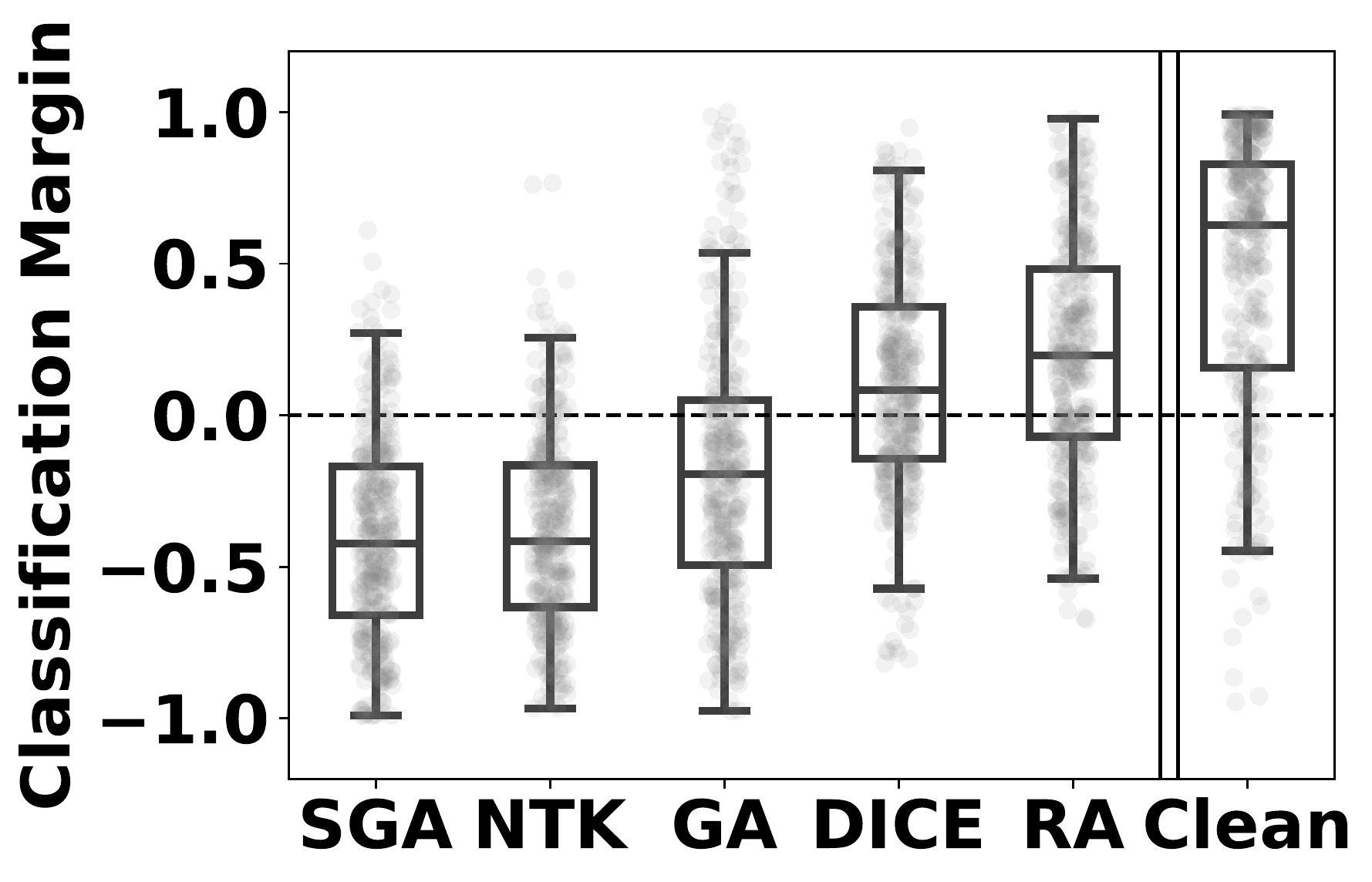}}\hspace{1mm}
  \subfigure[GraphSAGE]
  {\includegraphics[width=0.185\linewidth]{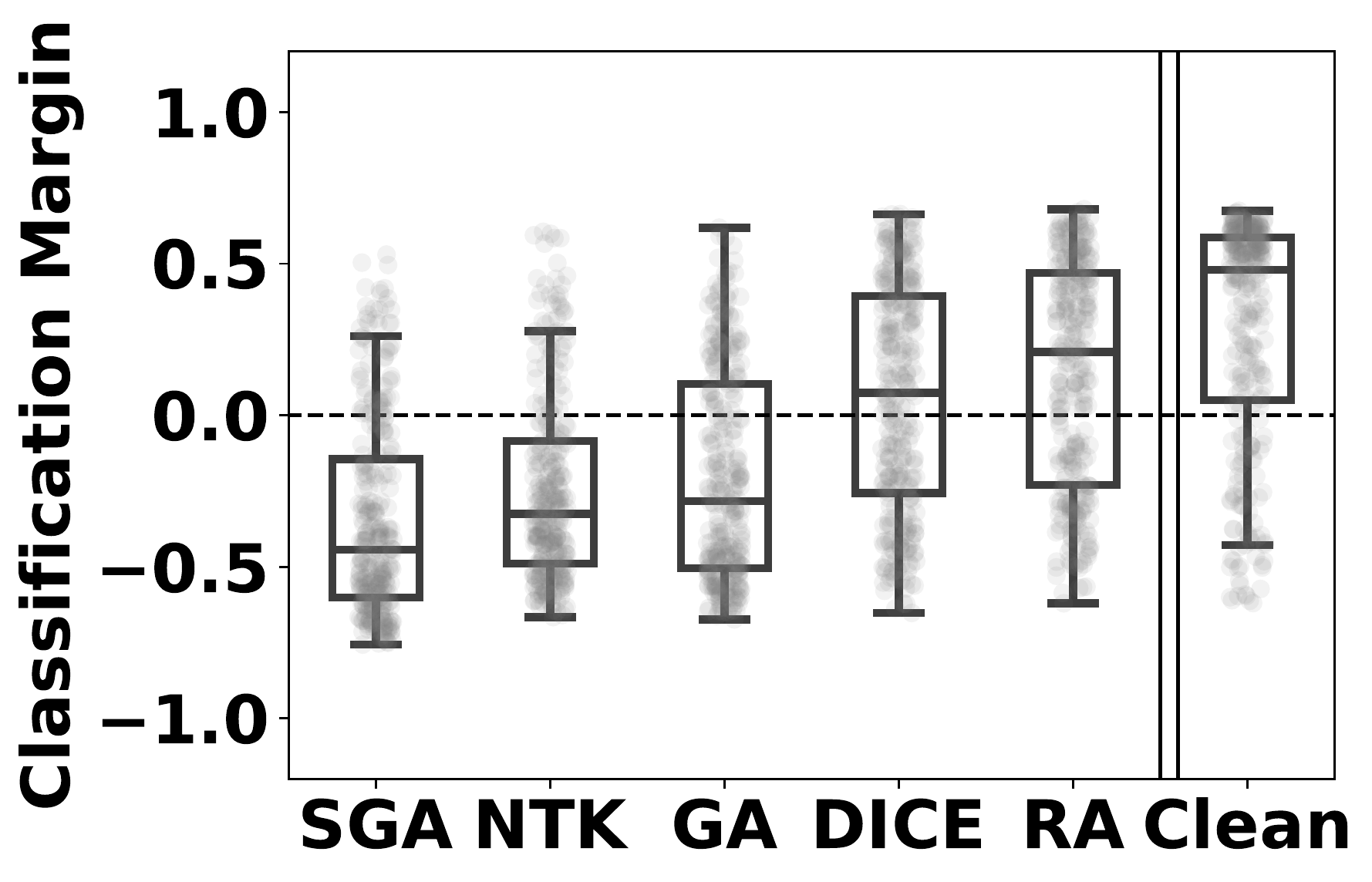}}\hspace{1mm}
  \subfigure[Cluster-GCN] {\includegraphics[width=0.185\linewidth]{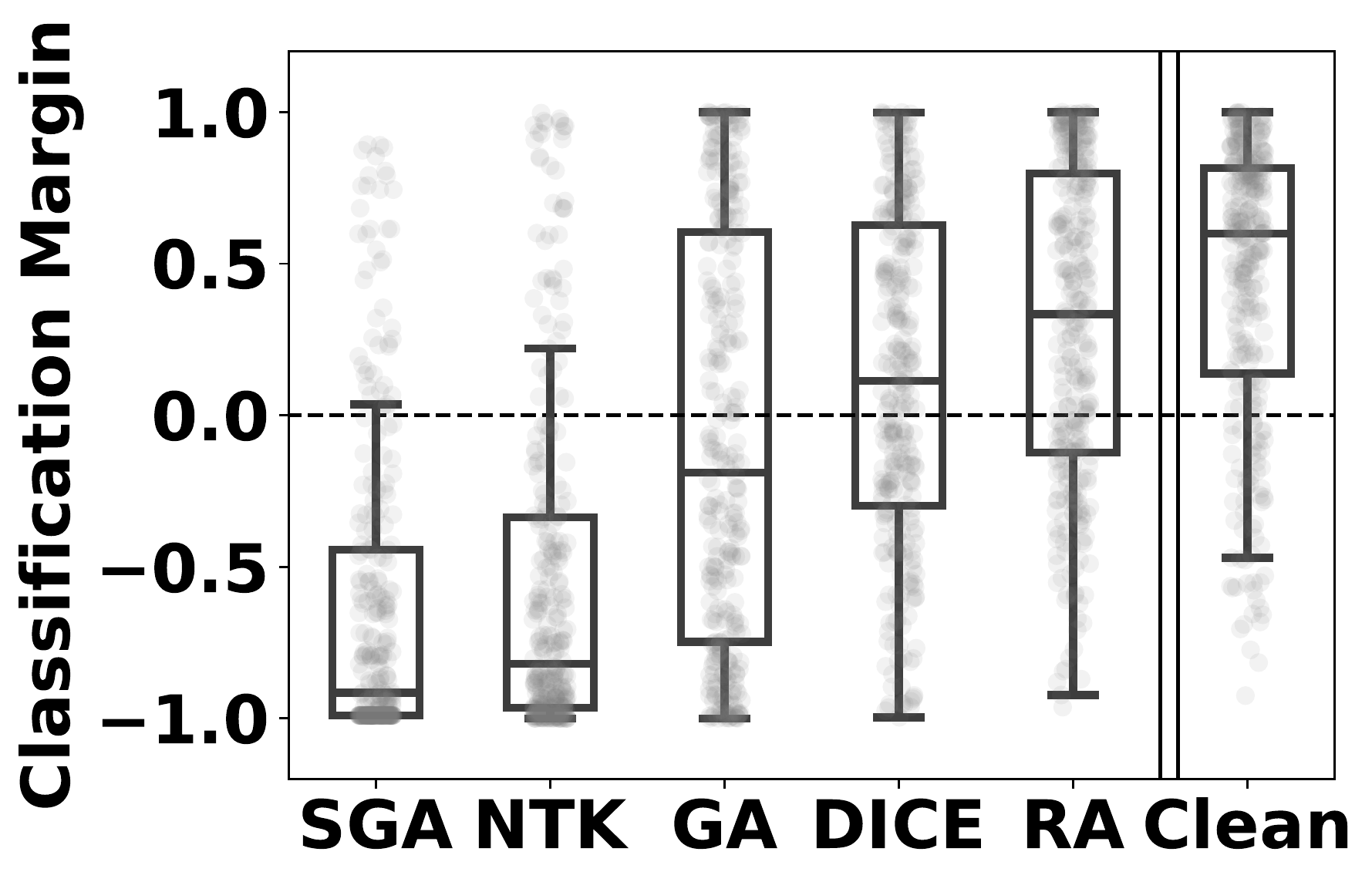}}\\

  \subfigure[SGC-In] {\includegraphics[width=0.185\linewidth]{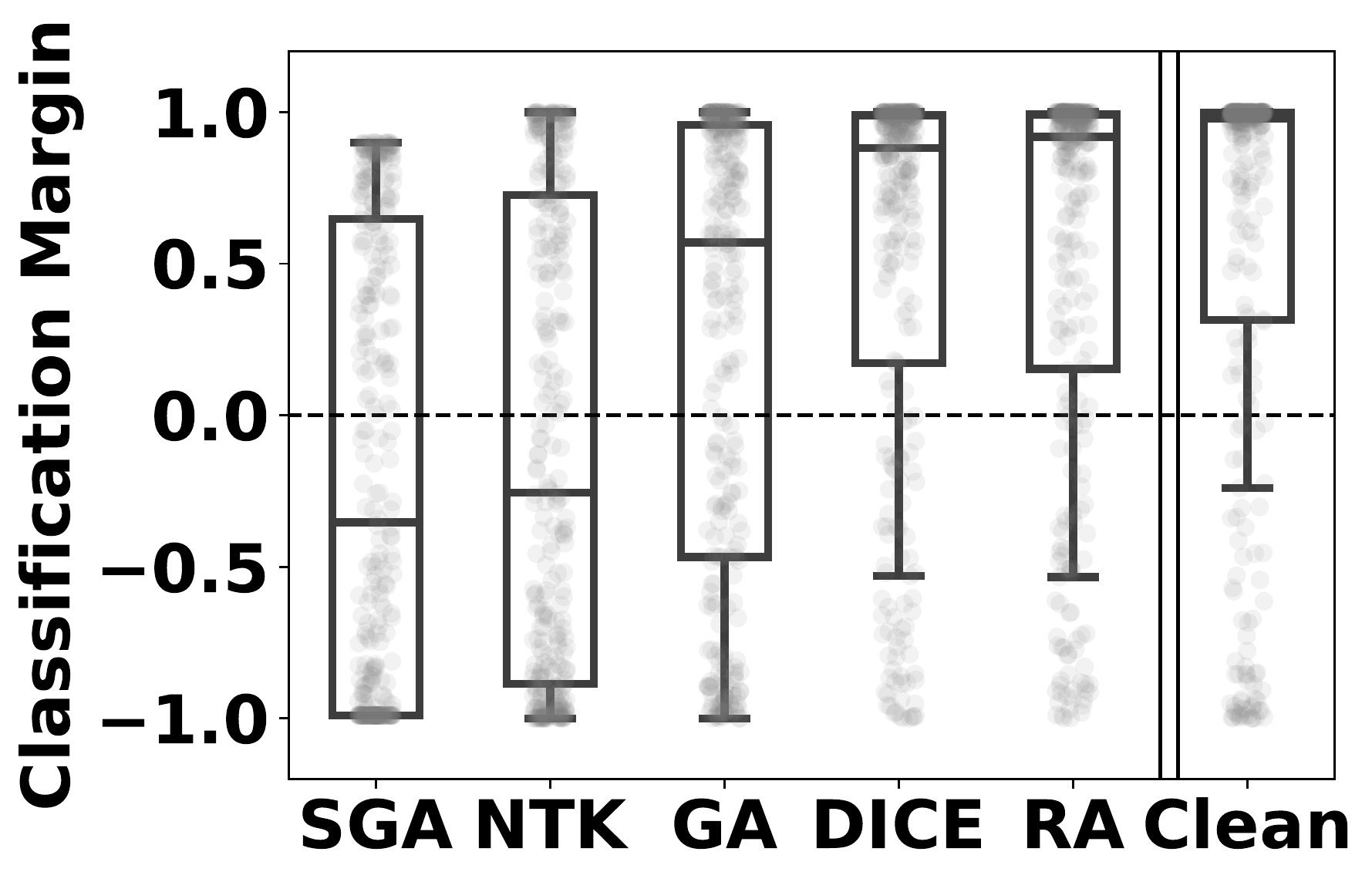}}\hspace{1mm}
  \subfigure[GCN-In] {\includegraphics[width=0.185\linewidth]{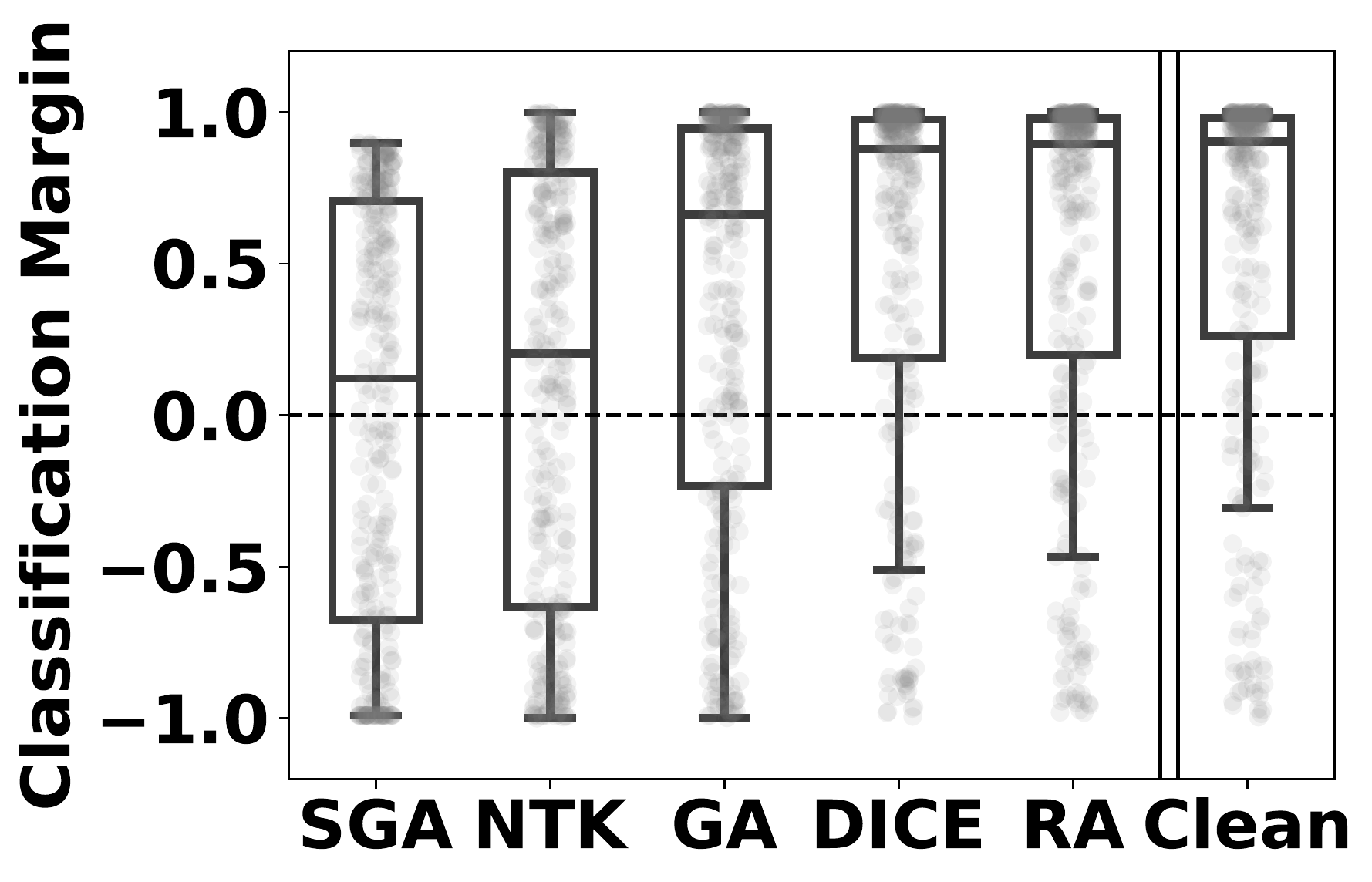}}\hspace{1mm}
  \subfigure[GAT-In] {\includegraphics[width=0.185\linewidth]{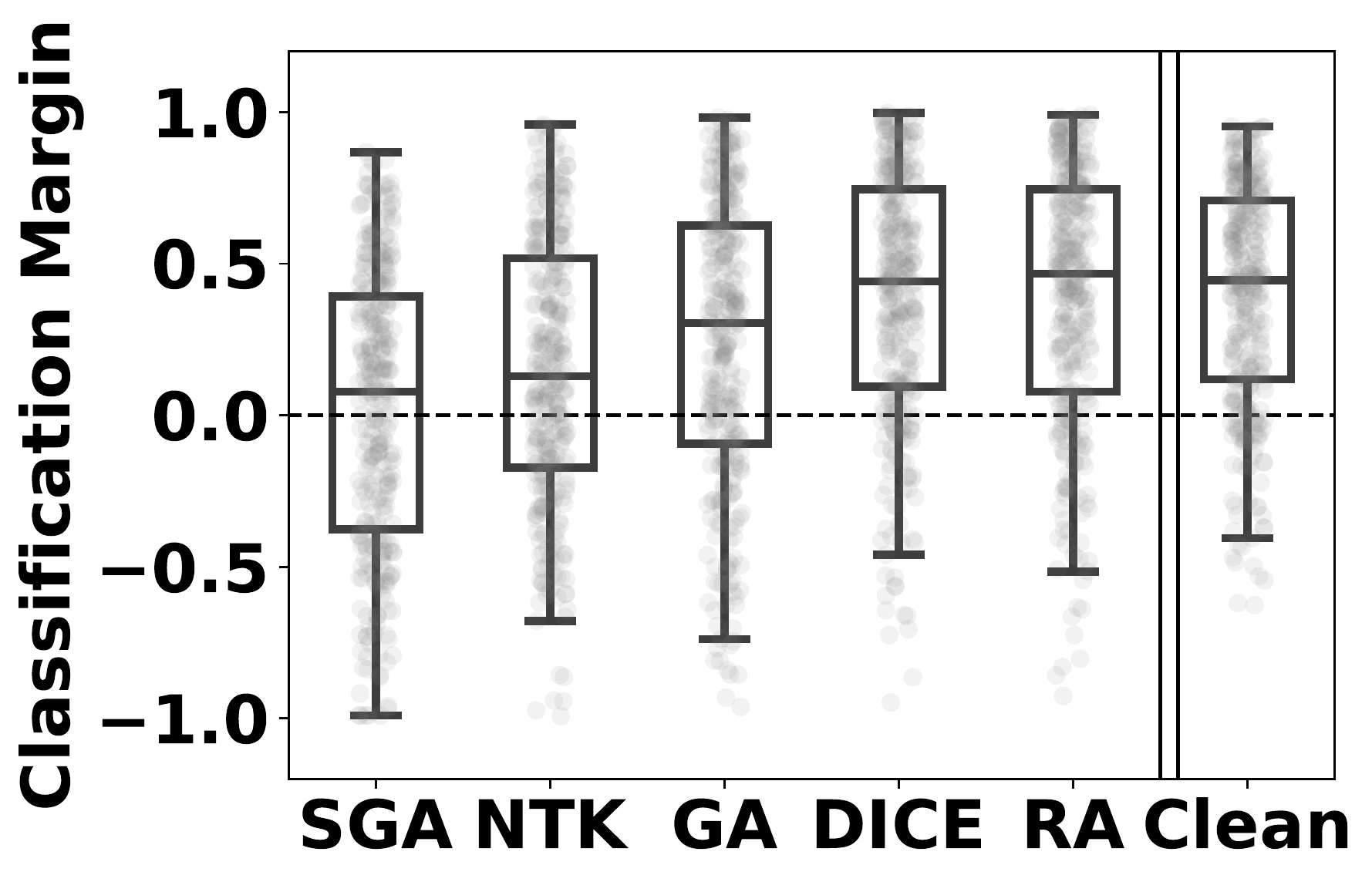}}\hspace{1mm}
  \subfigure[GraphSAGE-In] {\includegraphics[width=0.185\linewidth]{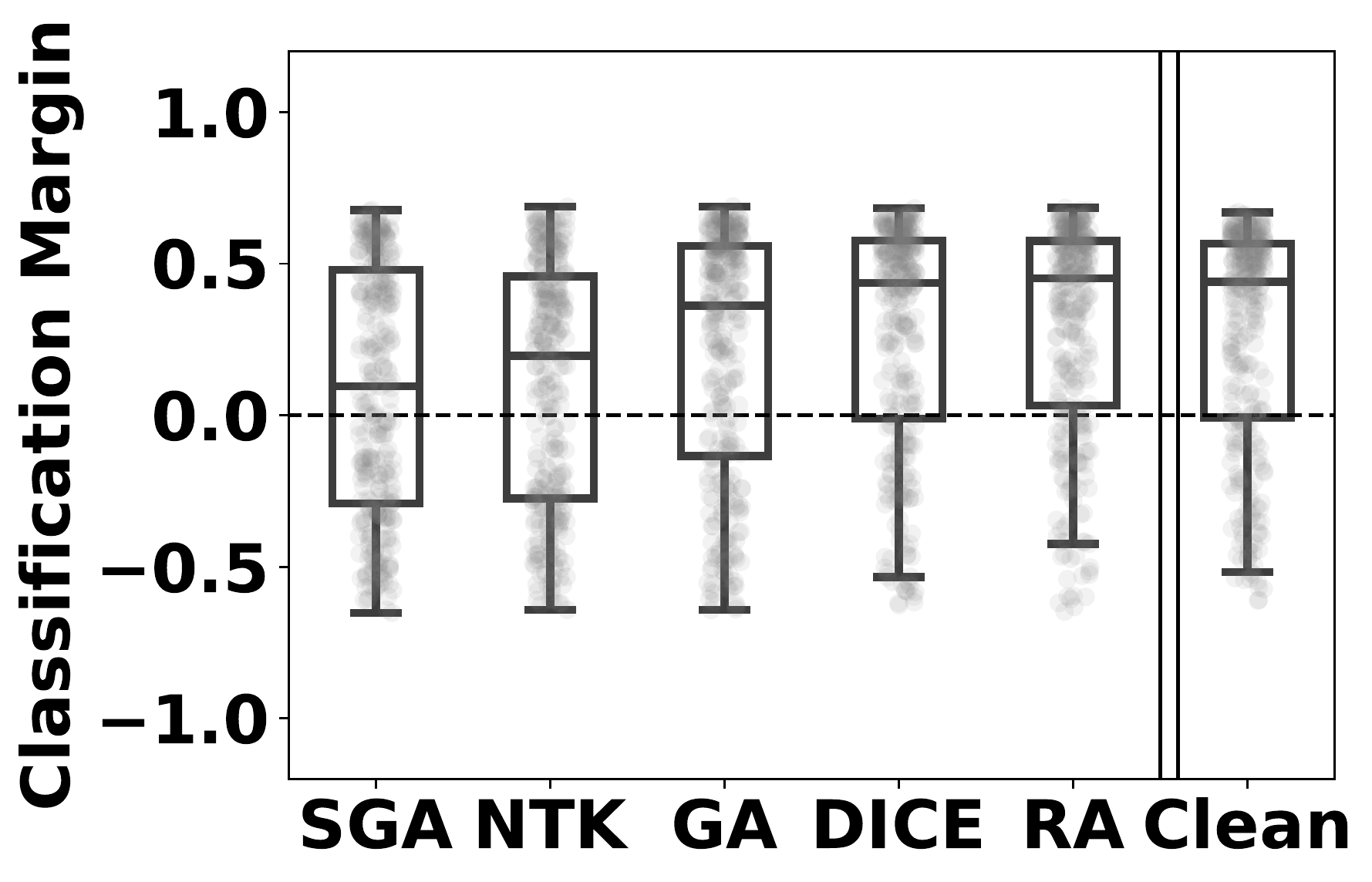}}\hspace{1mm}
  \subfigure[Cluster-GCN-In] {\includegraphics[width=0.185\linewidth]{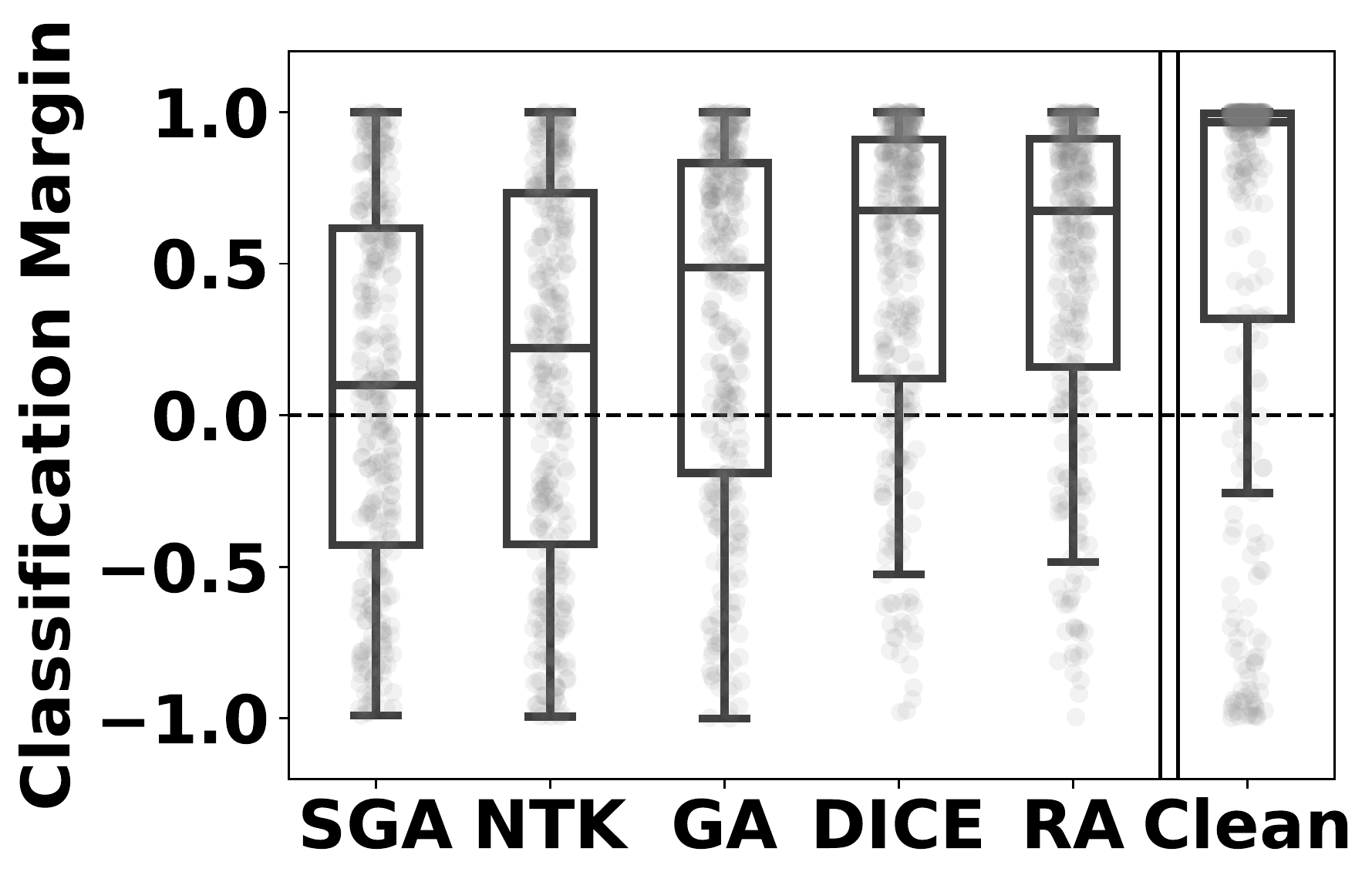}}
  \caption{Results on Cora dataset using direct attack (top) and influence attack (-In, bottom). ``NTK'' is short for Nettack and ``GA'' is short for GradArgmax.}
  \label{fig_margin}
\end{figure*}

\subsection{Performance on Generating Adversarial Examples (RQ1)}
As described in Table \ref{complexity}, our method is highly efficient in terms of time and space complexity compared to other methods. For running time, here we theoretically analyze how much improvement our method can achieve over Nettack, the state-of-the-art adversarial attack method. Following the parameter settings described in Section \ref{parm}, i.e., $k=2$ and $\Delta=d$, the relative time complexity can be approximated as
\begin{equation}
  \label{eq_times}
  \frac{\mathcal{O}(d \cdot |\mathcal{A}| \cdot N\cdot d^2)}{\mathcal{O}(d \cdot (d^2+d \cdot |\mathcal{A}|)\cdot d^2)}=\mathcal{O}(\frac{|\mathcal{A}| \cdot N}{d^2+d \cdot |\mathcal{A}|})\,.
\end{equation}
This means that SGA can theoretically achieve $\frac{|\mathcal{A}| \cdot N}{d^2+d \cdot |\mathcal{A}|}$ times speedup compared to Nettack, and it will be higher for larger and sparser graphs.

Table \ref{time_mem_DAC} shows the performance of generating adversarial examples with direct and influence attack strategies, respectively. Here the training time of the surrogate model is excluded and memory usage refers to the usage of storing the graph (subgraph). Please note that we do not compare the running time and memory usage with RA and DICE since they both are random algorithms and comparison doesn't make sense. Because the first three datasets are sparse enough with small degrees on average (see Table \ref{dataset}), the two-hop subgraph is much smaller and SGA has a remarkable improvement in efficiency. Intuitively, we plot the performance of Nettack and GradArgmax over their running time relative to that of SGA on three datasets in Figure \ref{fig_times}. Considering the Citeseer dataset (statistics are in Table \ref{dataset}) and the direct attack setting (i.e., $|\mathcal{A}|=1$), the theoretical speedup comes to 134 times according to Eq.(\ref{eq_times}), which is approximately consist with the experimental results (120 times).

Table \ref{time_mem_DAC} indicates that our method is much more efficient than GradArgmax and Nettack in terms of time and space complexity. Even if the dataset grows in size, SGA remains high efficiency as if the graph is sparse enough. On Pubmed dataset, our method can even yield up to three orders of magnitude speedup over Nettack. Also, SGA is much more memory efficient. On the contrary, Nettack becomes less efficient, especially when performing influence attacks on a larger dataset, the reason is that the larger scale of attacker nodes $\mathcal{A}$ and candidate edges set. As for GradArgmax, the running time and memory usage are similar between direct and influence attacks, since it computes the gradients of the whole graph all the time and the candidate edges set are always the same. In addition, both of them failed on Reddit dataset, a large and dense graph, but SGA still achieves high efficiency.

In respect to the metric DAC, Table \ref{time_mem_DAC} shows that: (i) Direct attacks have greater impacts than influence attacks, it is clear and interpretable since perturbations are restricted in the neighborhood of the target node, leading to a significant degree changes especially the target node. Naturally, the concentrated perturbations will exert a greater influence; (ii) Nettack achieves the most unnoticeable influence because it is enforced to preserve the graph's degree distribution during attacks. Our method leverages the subgraph instead of the whole graph without any constraints on the degree distribution, thus achieving a slightly worse result but still better than GradArgmax. Note that, GradArgmax considers all $N^2$ edges as a candidate set to modify, it will largely affect the degree distribution of nodes in the graph, and the attack becomes more noticeable.

% Note that, in the scenario of influence attacks, RA and DICE become slightly worse since they tend to affect more nodes in the graph, while others will concentrate on a part of nodes with high possibility to affect the target node.

\subsection{Performance on Attacking Graph Neural Networks (RQ2)}

\label{attackperformance}
For the target model SGC, whose details of the architecture are transparent (but not for its weights), it can be approximately treated as a white-box attack if it is used as a surrogate model. However, for other graph neural network models, this is a complete black-box attack without any prior knowledge.

As shown in Table \ref{attack_small}, we report the percentages of target nodes that are correctly classified, where \emph{clean} stands for results in the original graph. The classification performance of SGC drops significantly against direct attacks. SGC is more vulnerable than other classifiers against influence attacks. Unlike SGC, other models are more robust even in the direct attack setting. There are four possible reasons to explain this:
(i) The nonlinear activation function. SGC and GCN are similar except for activation functions. SGC drops the nonlinear activation in the hidden layer and collapses weight matrices between consecutive layers. Although SGC behaves more efficiently during training, the simplification leads to lower robustness than GCN.
(ii) The details of other models' architecture are not exposed to attackers like SGC, so the performance depends on the transferability of attacks;
(iii) The message aggregation methods are more robust than simple graph convolution. For instance, GAT introduces the attention mechanism and enables (implicitly) to specify different weights to different nodes in a neighborhood. For GraphSAGE, the most robust one in most cases, we argue that it is probably on account of the \emph{concatenate} operation on the node's message and the neighborhoods', which can be regarded as a residual term between layers. Note that we only consider structure attack rather than feature attack, the node feature of itself is unperturbed. By doing so, the influence of attacker nodes will be largely alleviated.
(iv) Pre-processing on the input graph. For Cluster-GCN, it samples several subgraphs whose nodes have high correlations using a graph clustering algorithm and restricts the message aggregation within these subgraphs, which also alleviates the influence of attacker nodes.

In Figure \ref{fig_margin}, we can observe that SGA achieves a considerable performance in most cases compared with different attack methods. Nettack, a strong baseline, also yields a significant performance as reported in \cite{zugner2018adversarial}. Most remarkably, even in attacking other robust graph neural networks (GAT, GraphSAGE, Cluster-GCN), most of the classifiers are strongly affected by perturbations especially performed by SGA and Nettack. The obtained results have proved the vulnerability of graph neural networks and are consistent with the previous studies. Not surprisingly, influence attacks achieve a worse performance compared with direct attacks. As expected, random algorithms RA and DICE both have a slight attack effect on attacking different graph neural networks. Both GradArgmax and SGA are gradient-based methods, we can also see that GradArgmax performs worse than our method SGA although the whole graph is used to attack. The possible explanations for the results are as follows: (i) The misclassification loss of SGA considers the loss of the next most probable class label as well as the ground-truth label, which is available to better exploit the vulnerability of graph neural networks. (ii) As detailed in Section \ref{sec:gradient}, the surrogate model appears to make an overconfident output and it causes the vanishing of gradients. SGA introduces a scale factor to calibrate the surrogate model, the gradient vanishing is alleviated and thus achieves a better performance. (iii) SGA focuses on a part of nodes --- the $k$-hop subgraph centered as the target node with some potential nodes. As a result, the generated perturbations will be more concentrated and cause the misclassification of target classifiers much easier.

\begin{table}[t]
  \centering
  \caption{Accuracy(\%) of direct attack and influence attack on two large scale datasets. Here the best results are \textbf{boldfaced}.}
  \label{attack_large}
  \resizebox{\linewidth}{!}{
    \begin{tabular}{c|ccccc|cc}
      % \toprule
      \hline
      Methods          & \multicolumn{5}{c|}{Pubmed} & \multicolumn{2}{c}{Reddit}                                                                                 \\
      \hline
      \hline
      Direct Attack    & SGC                         & GCN                        & GAT           & GraphSAGE     & Cluster-GCN   & SGC           & GCN           \\
      % \midrule
      \hline
      Clean            & 84.2                        & 86.5                       & 85.3          & 86.1          & 85.7          & 93.8          & 93.6          \\
      RA               & 76.6                        & 78.2                       & 75.0          & 81.5          & 75.8          & 88.6          & 89.4          \\
      DICE             & 65.0                        & 67.1                       & 57.8          & 72.3          & 64.0          & 84.2          & 88.2          \\
      GradArgmax       & 14.8                        & 15.8                       & 16.9          & 42.7          & 48.3          & -             & -             \\
      Nettack          & 2.0                         & 3.4                        & 10.9          & 32.4          & 19.9          & -             & -             \\

      % \midrule
      \hline
      SGA              & \textbf{1.6}                & \textbf{2.0}               & \textbf{9.3}  & \textbf{30.6} & \textbf{15.7} & \textbf{1.2}  & \textbf{5.4}  \\
      % \bottomrule
      % \toprule
      \hline
      \hline
      Influence Attack & SGC                         & GCN                        & GAT           & GraphSAGE     & Cluster-GCN   & SGC           & GCN\cr
      % \midrule
      \hline
      RA               & 84.1                        & 86.2                       & 85.3          & 87.2          & 84.6          & 93.6          & 93.5          \\
      DICE             & 83.6                        & 85.9                       & 84.7          & 85.8          & 84.4          & 93.7          & 93.7          \\
      GradArgmax       & 73.2                        & 77.8                       & 72.7          & 84.0          & 76.2          & -             & -             \\
      Nettack          & 61.4                        & 72.0                       & 70.1          & 85.4          & 69.2          & -             & -             \\

      % \midrule
      \hline
      SGA              & \textbf{63.4}               & \textbf{70.7}              & \textbf{70.0} & \textbf{82.6} & \textbf{67.8} & \textbf{78.9} & \textbf{83.6} \\
      % \bottomrule    
      \hline
    \end{tabular}
  }
\end{table}

\subsection {Scalability for Larger Datasets (RQ3)}
As illustrated above, SGA achieves a considerable performance on attacking most graph neural networks on two small-scale datasets Cora and Citeseer, and SGA also ensures time and space efficiency. To evaluate the scalability of SGA, we extend our experiments to two larger datasets --- Pubmed and Reddit, the data statistics are described in Table \ref{dataset}. Particularly, Reddit is a relatively dense graph with node degrees up to $99.65$ on average, which means that the attack budgets are much higher than other datasets in our settings, and it naturally brings more challenges on time and memory usage.

As shown in Table \ref{time_mem_DAC}, GradArgmax and Nettack become slower, and more memory usage is required on larger datasets. Especially for the Reddit dataset, both GradArgmax and Nettack have failed due to high time and space complexity. On the contrary, SGA has similar time and memory usage on Pubmed dataset as it is sparse as Citeseer and Cora. Even on Reddit dataset, SGA has high efficiency in the direct attack (12.236s) and the influence attack (14.571s) settings, respectively.

Furthermore, we conduct attacks on the same target models by the generated adversarial examples. As shown in Table \ref{attack_large}, SGA achieves state-of-the-art results in all cases, a significant performance decrease is observed on most of the target models especially SGC. As the data scale grows, random algorithms RA and DICE have little or no effect on target models. On the largest dataset Reddit, GradArgmax and Nettack have failed to conduct attacks. But for SGA, the obtained results show that only a small part of target nodes are correctly classified by SGC and GCN in the direct attack setting. In the influence attack setting, SGA also has an obvious effect on the target models compared to RA and DICE. Results on Pubmed and Reddit datasets suggest that our method can easily scale to larger datasets and significantly degrade the performance of target models.

\begin{figure}
  \centering
  \subfigure[Citeseer] {\includegraphics[width=0.45\linewidth]{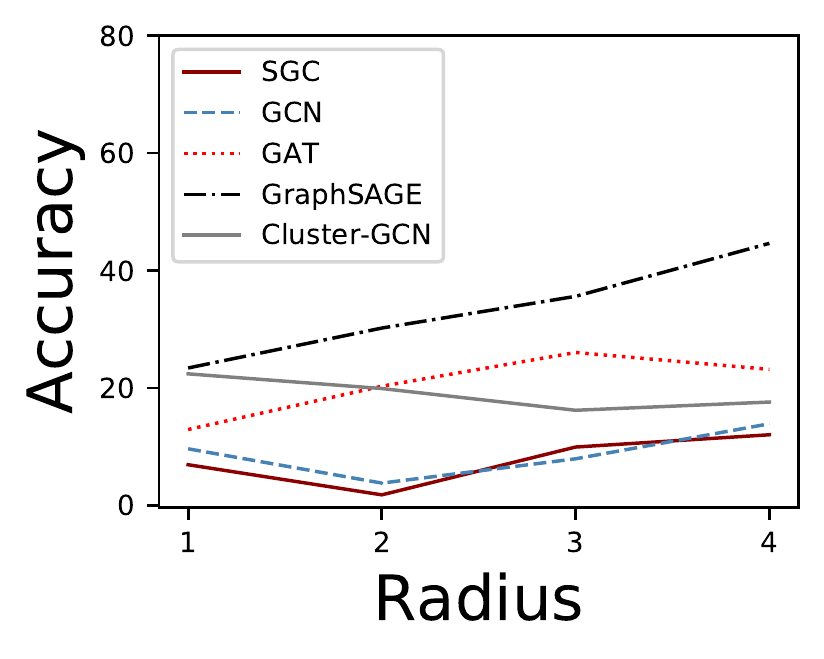}}\hspace{1mm}
  \subfigure[Cora] {\includegraphics[width=0.45\linewidth]{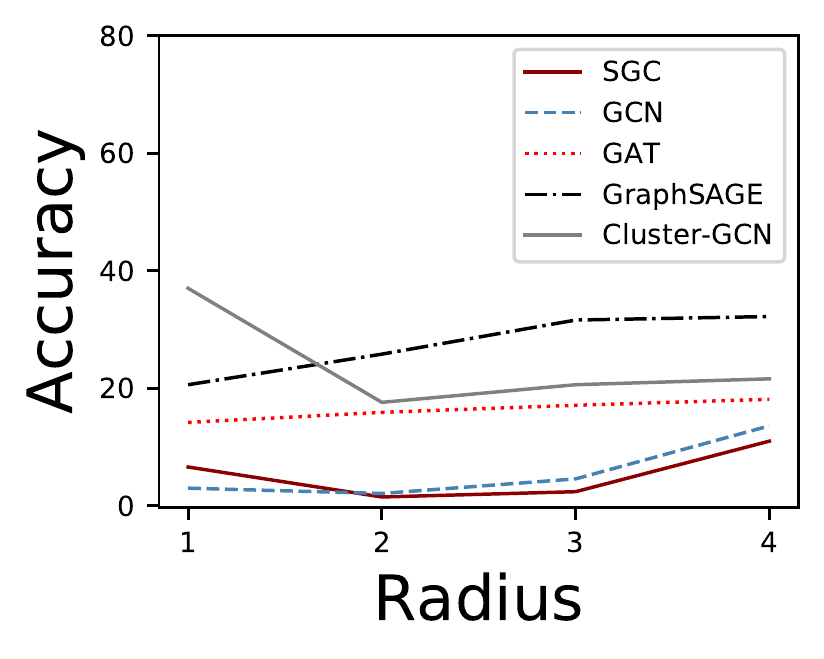}}\hspace{1mm}
  \caption{Performance of SGA on attacking different graph neural network models with various radius on Citeseer and Cora datasets.}
  \label{hops}
\end{figure}

\subsection{Ablation Study}

In order to explore the effect of different radius on SGA, we report the ablation test on Citeseer and Cora datasets from the various radius of the subgraph, i.e., $k=1,2,3,4$. In Figure \ref{hops}, it can be observed that $k=2$ achieves the best performance on attacking SGC and GCN on both datasets. But for other target models, a better performance is achieved when $k=1$ or $k=3$.

However, if $k=1$ the subgraph includes only the first-order neighboring nodes of the target node (except for potential nodes), SGA can only delete the edges that are directly connected to the target node but it is not allowed in the influence attack setting. As the radius of the subgraph gets larger, the number of nodes and edges increases exponentially, and it would require more time and memory usage to perturb the graph. So we need to make a trade-off between efficiency and performance. Given that the best classification performance can be achieved by a two-layer GCN or SGC in most datasets, SGA can exploit the vulnerability of graph neural network models with only a two-hops subgraph, and the efficiency of attack is also preserved.

\section{Discussions}
In this section, we will make a discussion about our proposed methods: SGA and DAC. Specifically, we detail how the proposed SGA can be generalized to other types of attack, i.e., feature attack and node injection attack. Besides, we discuss how to better ensure the unnoticeability of attackers by adopting the DAC metric during attacks.

\subsection{\emph{Feature Attack}}
SGA was initially designed for adversarial structure attack, it can be easily generalized to the feature attack (either binary or continuous features) scenario as well. Feature attack is simpler than structure attack since the structure of the subgraph is fixed and there is no need to update the subgraph after an attack. Specifically, there are two cases: (i) binary node features. This is the most typical case. For example, in a typical citation graph, nodes are documents, edges are citation links, and node features are bag-of-words feature vectors. It is easy to conduct feature attacks, just by taking into account the gradients of the input features. Similar to structure attack, we can choose to add or remove the feature based on the surrogate gradients. (ii) continuous node features. This is another case where node features can be derived from other node embedding methods (e.g., DeepWalk \cite{perozzi2014deepwalk} and Node2Vec \cite{grover2016node2vec}). We can treat it as continuous data by adopting gradient ascent to attack. This case is not suitable for our experimental settings since it is difficult to constrain such feature attacks within a given budget $\Delta \in\mathbb{N}$.

\subsection{\emph{Node Injection Attack}}
Node injection attack is a new attack scenario \cite{WangLSLYZ20}, where attackers can add a group of vicious nodes into the graph so that topological of the original graph can be avoided and the victim node will be misclassified. SGA is not available for such an attack. The injected vicious node can be regarded as a singleton node because it isn't initially connected to any nodes. Therefore, the strategy of extracting a $k$-hop subgraph doesn’t work in this case. However, the gradient-based method can be adapted to this scenario, which can be implemented by computing the gradients of all the non-edges between the injected node and other benign ones. Note that the computation still has a high time and memory usage as the graph scale grows. In the future, we plan to investigate more ways to improve the efficiency of conducting the node injection attack.

% How to conduct an efficient node injection attack is another challenge, we leave it as our future work.

\subsection{\emph{Unnoticeability Constraint}}
DAC is proposed as a practical metric to show the intensity of the impact after perturbations being performed. Although DAC overcomes the disadvantages of the traditional metrics, there is room for improvement. Specifically, the evaluation of attack impacts via DAC has occurred after attacks, while an attacker may prefer to adopting it as an unnoticeability constraint during attacks. For example, the results of DAC can be used as an additional loss of the attack. However, the computation of DAC is costly, which motivates us to improve it in future work.

% It is based on the network property and the principle cannot be applied to the node features. 

\section{CONCLUSION AND FUTURE WORK}
In this work, we study the adversarial attacks on the (attributed) graph and present a novel simplified gradient-based attack framework targeted on the node classification task. Specifically, we aim to poison the input graph and lead to the misclassification of graph neural networks on several target nodes. Based on the extensive experiments, our method SGA, which simply leverages a $k$-hop subgraph of the target node, has achieved high efficiency in terms of time and
space and also obtained competitive performance in attacking different models compared to other state-of-the-art adversarial attacks. It can be also observed that SGA scales to larger datasets and achieves a remarkable attack performance. In addition, we emphasize the importance of measuring the attack impacts on graph data and further propose DAC as a measure. The results show that DAC works as a practical metric and SGA can also achieve a relatively unnoticeable attack impact.

% 	Even though our method has achieved great performance, it generates adversarial examples without any restrict, therefore DAC are suggested to be taken into consideration in future. Moreover, 
Our works mainly focus on the node classification task and the targeted attack scenario. In future work, we aim to extend our method and generalize it to other graph analysis tasks with more flexibility.
% 	Studying the robustness of deep learning models is critical due to their vulnerability, we will focus more on enhancing the robustness of them in the future.

\ifCLASSOPTIONcompsoc
  \section*{Acknowledgments}
\else
  \section*{Acknowledgment}
\fi

The research is supported by the Key-Area Research and Development Program of Guangdong Province (No. 2020B010165003), the National Natural Science Foundation of China (No. U1711267),  the Guangdong Basic and Applied Basic Research Foundation (No. 2020A1515010831), and the Program for Guangdong Introducing Innovative and Entrepreneurial Teams (No. 2017ZT07X355).

\ifCLASSOPTIONcaptionsoff
  \newpage
\fi

\bibliographystyle{IEEEtran}
\bibliography{main}


%\bibliographystyle{IEEEtran}


%\newpage

\begin{IEEEbiography}[{\includegraphics[width=1in,height=1.25in,clip,keepaspectratio]{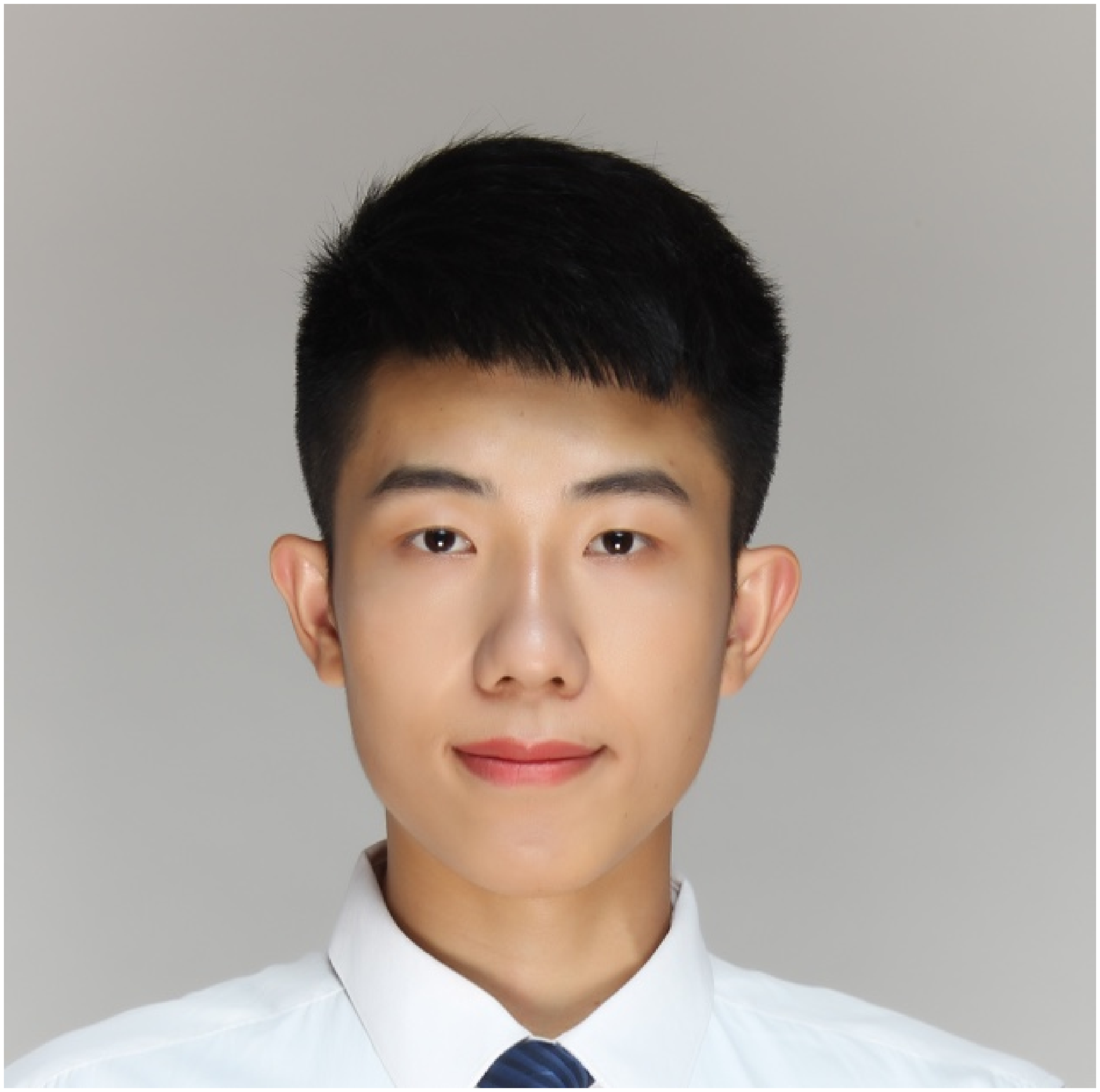}}]{Jintang Li}
  received the bachelor's degree at Guangzhou Medical University, Guangzhou, China, in 2018. He is currently pursuing the master's degree with the School of Electronics and Communication Engineering, Sun Yat-sen University, Guangzhou, China. His main research interests include graph representation learning, adversarial machine learning, and data mining techniques.
\end{IEEEbiography}

\begin{IEEEbiography}[{\includegraphics[width=1in,height=1.25in,clip,keepaspectratio]{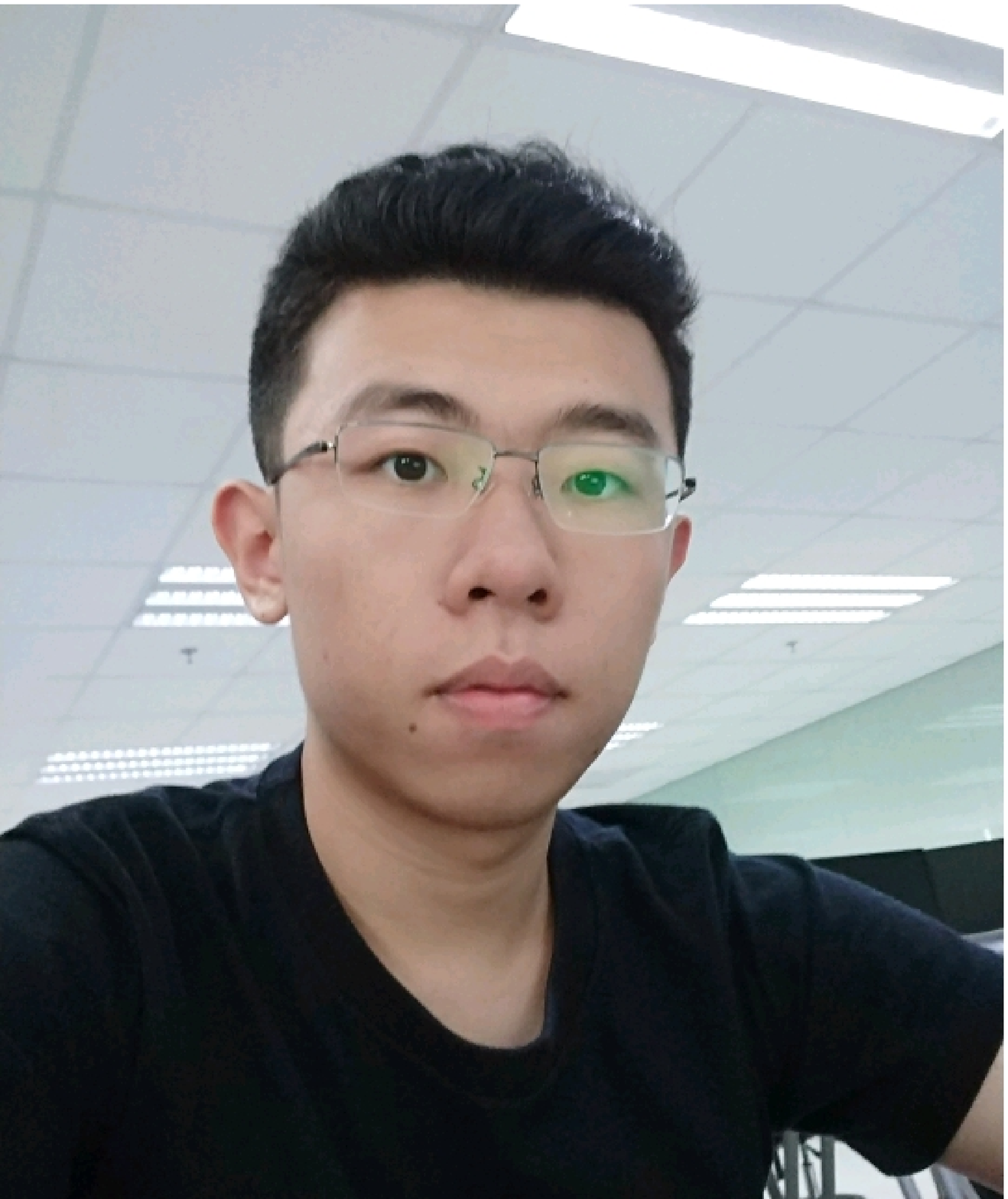}}]{Tao Xie}
  received the bachelor's degree at Sun Yat-sen University, Guangzhou, China, in 2020. He is currently pursuing the master's degree with the School of Computer Science and Engineering, Sun Yat-sen University, Guangzhou, China. His main research interests include recommendation systems, machine learning and data mining techniques.
\end{IEEEbiography}

\begin{IEEEbiography}[{\includegraphics[width=1in,height=1.25in,clip,keepaspectratio]{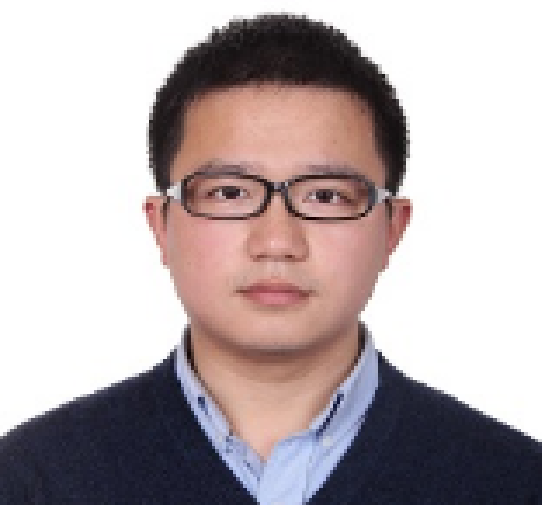}}]{Liang Chen}
  received the bachelor’s and Ph.D. degrees from Zhejiang University (ZJU) in 2009 and 2015, respectively. He is currently an associate professor with the School of Computer Science and Engineering, Sun Yat-Sen University (SYSU), China. His research areas include data mining, graph neural network, adversarial learning, and services computing. In the recent five years, he has published over 70 papers in sev- eral top conferences/journals, including SIGIR, KDD, ICDE, WWW, ICML, IJCAI, ICSOC, WSDM, TKDE, TSC, TOIT, and TII. His work on service recommendation has received the Best Paper Award Nomination in ICSOC 2016. Moreover, he has served as PC member of several top conferences including SIGIR, WWW, IJCAI, WSDM etc., and the regular reviewer for journals including TKDE, TNNLS, TSC, etc.
\end{IEEEbiography}

\begin{IEEEbiography}
  [{\includegraphics[width=1in,height=1.25in,clip,keepaspectratio]{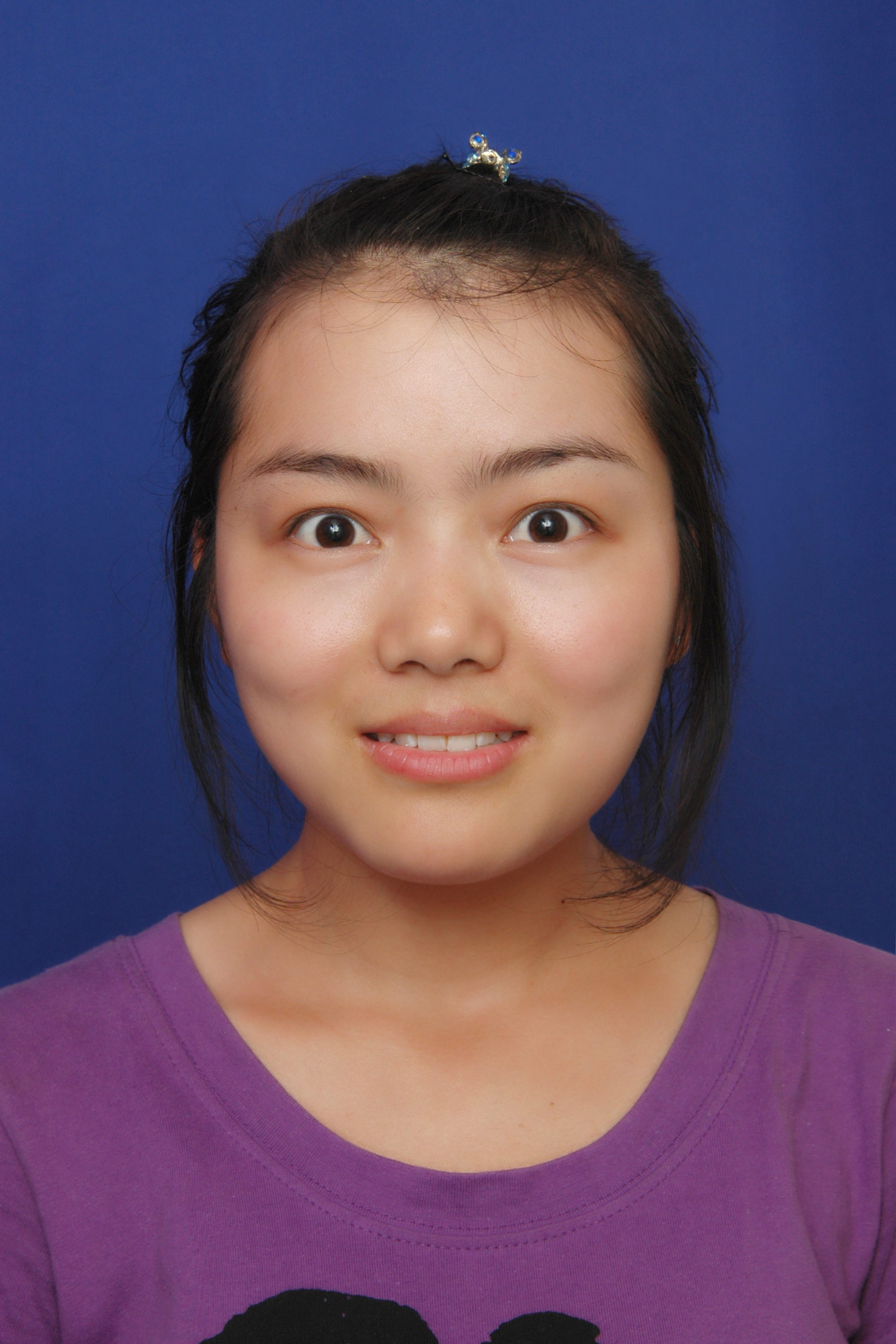}}]{Fenfang Xie} received the PhD degree in computer science and technology from Sun Yat-sen University, Guangzhou, China, in 2019. She is currently a postdoctoral fellow at the School of Computer Science and Engineering, Sun Yat-sen University, Guangzhou, China. Her research interests include recommender system, services computing, software engineering, machine learning and deep neural network.
\end{IEEEbiography}

\begin{IEEEbiography}[{\includegraphics[width=1in,height=1.25in,clip,keepaspectratio]{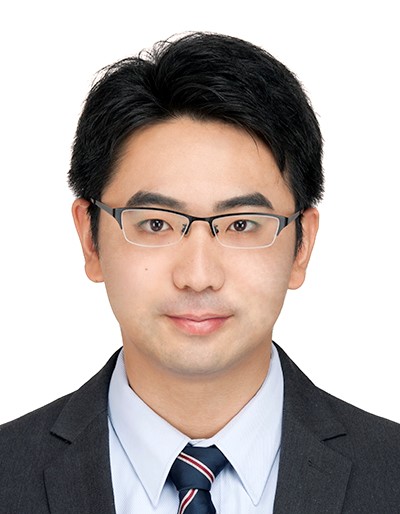}}]{Xiangnan He} received the Ph.D. in Computer Science from the National University of Singapore (NUS) in 2016. His research interests span information retrieval, data mining, and multi-media analytics. He has over 70 publications that appeared in several top conferences such as SIGIR, WWW, and MM, and journals including TKDE, TOIS, and TMM. His work on recommender systems has received the Best Paper Award Honorable Mention in WWW 2018 and ACM SIGIR 2016. Moreover, he has served as the PC chair of CCIS 2019, area chair of MM (2019, 2020) ECML-PKDD 2020, and PC member for several top conferences including SIGIR, WWW, KDD, WSDM etc., and the regular reviewer for journals including TKDE, TOIS, TMM, etc.
\end{IEEEbiography}

\begin{IEEEbiography}[{\includegraphics[width=1in,height=1.25in,clip,keepaspectratio]{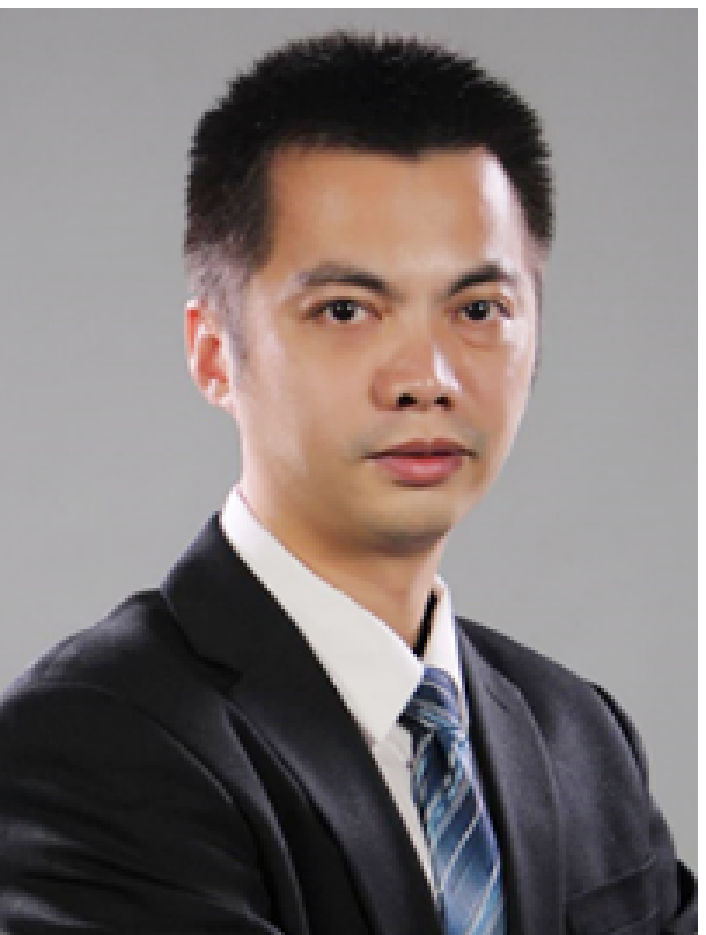}}]{Zibin Zheng}
  received the Ph.D. degree from The Chinese University of Hong Kong, in 2011. He is currently a Professor with the School of Computer Science and Engineering, Sun Yat-sen University, Guangzhou, China. His research interests include services computing, software engineering, and blockchain. He received the ACM SIGSOFT Distinguished Paper Award at the ICSE’10, the Best Student Paper Award at the ICWS’10, and the IBM Ph.D. Fellowship Award.
\end{IEEEbiography}

\end{document}